\documentclass[twoside]{article}

\usepackage{PRIMEarxiv}

\usepackage{hyperref}       
\usepackage{url}            
\usepackage{booktabs}       
\usepackage{amsfonts}       
\usepackage{nicefrac}       
\usepackage{microtype}      
\usepackage{lipsum}
\usepackage{fancyhdr}       
\usepackage{graphicx}       
\graphicspath{{media/}}     

\usepackage{xspace}
\usepackage[T1]{fontenc}
\usepackage[utf8]{inputenc}
\usepackage[portuguese]{babel}
\usepackage{multirow}
{}
{}
{}
\usepackage{booktabs}

\usepackage{balance}

\usepackage[labelformat=simple]{subcaption}

\usepackage[acronym]{glossaries}

\newacronymstyle{long-short-br}
{%
  \GlsUseAcrEntryDispStyle{long-short}%
}%
{%
  \GlsUseAcrStyleDefs{long-short}%
}

\title{ORCNet: A context-based network to simultaneously segment the ocular region components
}

\author{
  Diego Rafael Lucio, Luiz~A. Zanlorensi \\
  DINF  \\
  Federal University of Paraná \\
  Curitiba, Paraná, Brazil\\
  \texttt{\{drlucio, lazjunior\}inf.ufpr.br} \\
   \And
   Yandre Maldonado e Gomes da Costa \\
   PCC/DIN \\
   State University of Maringá  \\
   Maringá, Paraná, Brazil \\
   \texttt{yandre@din.uem.br} \\
   \AND
   David Menotti \\
   DINF  \\
  Federal University of Paraná \\
  Curitiba, Paraná, Brazil\\
  \texttt{menotti@inf.ufpr.br} \\
}

\begin{document}
\maketitle

\begin{abstract}
Accurate extraction of the Region of Interest is critical for successful ocular region-based biometrics. In this direction, we propose a new context-based segmentation approach, entitled Ocular Region Context Network (ORCNet), introducing a specific loss function, i.e., the Punish Context Loss (PC-Loss). 
The PC-Loss punishes the segmentation losses of a network by using a percentage difference value between the ground truth and the segmented masks. 
We obtain the percentage difference by taking into account Biederman’s semantic relationship concepts, in which we use three contexts (semantic, spatial, and scale) to evaluate the relationships of the objects in an image. Our proposal achieved promising results in the evaluated scenarios --- iris, sclera, and ALL (iris + sclera) segmentations ---, outperforming the literature baseline techniques.
The ORCNet with ResNet-152 outperforms the best baseline (EncNet with ResNet-152) on average by $2$.$27\%$,  $28$.$26\%$ and $6$.$43\%$ in terms of F-Score, Error Rate and Intersection Over Union, respectively. We also provide (for research purposes) $3$,$191$ manually labeled masks for the MICHE-I database, as another contribution of our work.
\end{abstract}



\newacronym{ar}{AR}{AR}
\newacronym{asm}{ASM}{Active Shape Model}
\newacronym{asef}{ASEF}{Average of Synthetic Exact Filters}
\newacronym{leasm}{LE-ASM}{Local Eyebrow Active Shape Model}

\newacronym{bovw}{BOVW}{Bag of Visual Words}
\newacronym{cln}{CLN}{Context Learning Network}
\newacronym{cnn}{CNN}{Convolutional Neural Network}
\newacronym{crf}{CRF}{Conditional Random Field}
\newacronym{cmos}{CMOS}{Complementary Metal-Oxide-Semiconductor}
\newacronym{encdec}{ED}{Encoder-Decoder}
\newacronym{encnet}{EncNet}{Context Encoding Network}
\newacronym{eer}{ER}{Error Rate}
\newacronym{eer2}{EER}{Equal Error Rate}
\newacronym{ercnet}{ORCNet}{Ocular Region Context Network}
\newacronym{fcn}{FCN}{Fully Convolutional Network}
\newacronym{gac}{GAC}{Geodesic Active Contours}
\newacronym{gan}{GAN}{Generative Adversarial Network}
\newacronym{gpu}{GPU}{Graphical Processor Unit}
\newacronym{hcnn}{HCNN}{Hierarchical Convolutional Neural Network}
\newacronym{hog}{HOG}{Histogram of Oriented Gradients}
\newacronym{iou}{IoU}{Intersection over Union}
\newacronym{iupui}{IUPUI}{IUPUI Multiwavelength}
\newacronym{map}{mAP}{mean Average Precision}
\newacronym{masdv1}{MASD.v1}{Multi-Angle Sclera Dataset.v1}
\newacronym{mer}{MER}{Mean Error Rate}
\newacronym{mfcn}{MFCN}{Multi-scale Fully Convolutional Network}
\newacronym{mlp}{MLP}{Multi-Layer Perceptron}
\newacronym{mtm}{MTM}{Markovian Texture Model}
\newacronym{mrs}{MRS}{Maximum Radial Suppression}
\newacronym{msrc}{MSRC}{Microsoft Research Cambridge}
\newacronym{nir}{NIR}{near-infrared}
\newacronym{ocac}{OCAC}{Optical Correlation based Active Contours}
\newacronym{prd}{PRD}{Periocular Region Detection}
\newacronym{pcloss}{PC-Loss}{Punish Context Loss}
\newacronym{roc}{ROC}{Receiver Operating Characteristics}
\newacronym{relu}{Leaky ReLU}{Leaky Rectified Linear Unit}
\newacronym{vfc}{VFC}{Vector Field Convolution}

\newacronym{roi}{RoI}{Region of Interest}
\newacronym{nice1}{NICE-I}{Noisy Iris Challenge Evaluation, Part I}
\newacronym{svm}{SVM}{Support Vector Machines}
\newacronym{vj}{VJ}{Viola-Jones}
\newacronym{vis}{VIS}{visible}
\newacronym{rpn}{RPN}{Region Proposal Network}

\newacronym{frr}{FRR}{False Reject Rate}

\newacronym{seloss}{SE-Loss}{Semantic Encoding Loss}

\newacronym{fp}{FP}{False Positive}
\newacronym{fn}{FN}{False Negative}

\newcommand{\distance}{CASIA-Iris-Distance\xspace}
\newcommand{\synth}{CASIA-Iris-Synt\xspace}
\newcommand{\warsaw}{LivDet-Iris 2015 Warsaw\xspace}
\newcommand{\clarkson}{LivDet-Iris 2015 Clarkson\xspace}
\newcommand{\berc}{BERC-Iris-Fake\xspace}

\newcommand{\ndiris}{ND-Iris-040\xspace}
\newcommand{\iitd}{IITD\xspace}
\newcommand{\wvu}{WVU Non-ideal\xspace}

\newcommand{\lamp}{CASIA-Iris-Lamp\xspace}
\newcommand{\resnet}{ResNet\xspace}
\newcommand{\ar}{AR\xspace}
\newcommand{\mobbio}{MobBIO\xspace}
\newcommand{\thousand}{CASIA-Iris-Thousand\xspace}
\newcommand{\ubiris}{UBIRIS.v2\xspace}
\newcommand{\ubirisi}{UBIRIS.v1\xspace}

\newcommand{\cross}{Cross-Eyed-VIS\xspace}
\newcommand{\csip}{CSIP\xspace}
\newcommand{\miche}{MICHE-I\xspace}
\newcommand{\niceii}{NICE-II\xspace}
\newcommand{\nicei}{NICE-I\xspace}
\newcommand{\polyu}{PolyU-VIS\xspace}
\newcommand{\visob}{VISOB\xspace}

\newcommand{\faster}{Faster~R-CNN\xspace}
\newcommand{\yolo}{YOLOv2\xspace}
\newcommand{\mobbiofake}{MobBIOfake\xspace}
\newcommand{\segnet}{SegNet\xspace}
\newcommand{\uninet}{UniNet.v2\xspace}
\newcommand{\unet}{U-Net\xspace}

\newacronym{fast}{Fast R-CNN}{Fast Region-Based Convolutional Neural Network}
\newacronym{fpn}{FPN}{Feature Pyramid Network}

\newacronym{gs4}{MICHE-GS4}{MICHE Galaxy S4 Subset}
\newacronym{gt2}{MICHE-GT2}{MICHE Galaxy Tab 2 Subset}
\newacronym{ip5}{MICHE-IP5}{MICHE Iphone 5 Subset}
\newacronym{masd}{MASD}{Multi-Angle Sclera Database}
\newacronym{gs4_1000}{MICHE1000-GS4}{MICHE 1000 Galaxy S4 Subset}
\newacronym{gt2_1000}{MICHE1000-GT2}{MICHE 1000 Galaxy Tab 2 Subset}
\newacronym{ip5_1000}{MICHE1000-IP5}{MICHE 1000 Iphone 5 Subset}

\newacronym{iiitdcli}{IIIT-D CLI}{IIIT-Delhi Contact Lens Iris}

\newacronym{ndcld15}{NDCLD15}{Notre Dame Contact Lens Detection 2015}

\newacronym{mobbio}{MobBIO}{MobBIO Subset}

\newacronym{ndccl}{NDCCL}{Notre Dame Cosmetic Contact Lenses}

\newacronym{frgc}{FRGC}{Face Recognition Grand Challenge}
\newacronym{mbgc}{MBGC}{Multiple Biometrics Grand Challenge}
\newacronym{mrf}{MRF}{Markov Random Field}

\vspace{10mm}

\section{Introduction}
\label{sec:introduction}

Recently, the interest in the use of biometrics to automatically verify or identify a person has increased. The fact that the biometric features (physical and behavioral) of a person cannot be lost or forgotten, as it can happen with other means of identification, such as passwords or identity cards, justifies this growth ~\cite{bolle2004guide,Bowyer2008}. 

Several physical characteristics of the human body, such as fingerprints, face, voice, and ocular region components, can be used as input for biometric systems. 
Among the previously mentioned physical attributes, the ocular region components have high discriminative power and, due to this fact, these components are an excellent choice to develop a non-invasive user identification system~\cite{das2014sclera,zhou2012new, Jain50years_2016, wildes1997iris, zanlorensi2018impact, LIU2016154, zanlorensi2019ocular}. 

\begin{figure}[!h]
	\begin{center}
		\includegraphics[width=.7\linewidth]{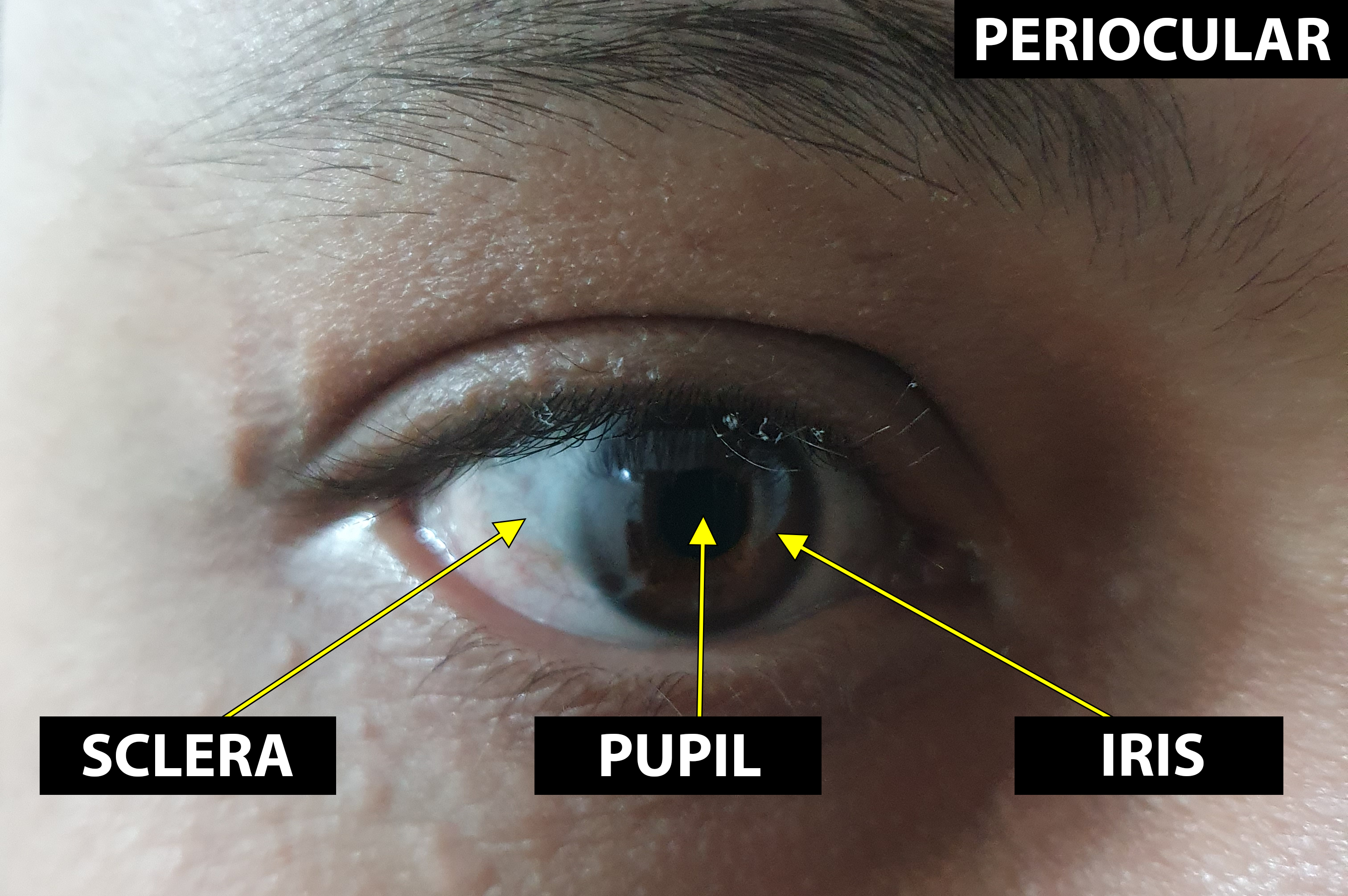}
		\caption{Ocular region components employed in biometric systems.}
		\label{fig:periocular_components}
	\end{center}
\end{figure}

Fig.~\ref{fig:periocular_components} shows the iris, pupil, sclera, and the periocular region, 
which are the most commonly employed ocular region components in biometrics.
According to Bowyer et al.~\cite{Bowyer2008}, the pupil region is the central portion of the eye and it is generally darker than the iris.
However, in some cases, there may be specular reflexes and cataracts, possibly making it clearer.
The iris is a colored ring composed of tissue placed around the pupil through which light enters the eye.
The sclera is a white region of connective tissue and blood vessels that surrounds the iris.
For the periocular region, there is no standard definition in the literature regarding the location of the \gls{roi}. 
Some researchers considered the center of the iris as a reference point and calculated the width and the height of \gls{roi} as $6\times$ and $4\times$ the radius of the iris, respectively ~\cite{mahalingam_2014,Park2011,Tan2013}. In contrast, Padole and Proença~\cite{Padole2012} proposed the use of the eyes corners as the reference point to calculate the \gls{roi} as they are less affected by gaze, pose variation and occlusion.

Among the previously mentioned ocular region components, the iris presents one of the most accurate results, since it has a sufficiently complex texture patterns that can be used on the identification task~\cite{LIU2016154, Al-Waisy2018, Nguyen_iris, proenca_irina}. However, in the last years, the experiments carried out using the sclera and the entire ocular region have also presented promising results on biometric tasks~\cite{LUZ20182, Proenca_Periocular2018, das2016ssrbc, das2017sserbc, delna2016sclera}.

\begin{figure*}[!h]
	\centering
	\begin{subfigure}[b]{0.18\textwidth}
		\includegraphics[width=\textwidth]{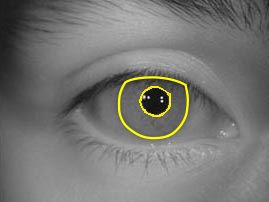}
		\caption{}
		\label{fig:circular_iris_detection}
	\end{subfigure}
	\begin{subfigure}[b]{0.18\textwidth}
		\includegraphics[width=\textwidth]{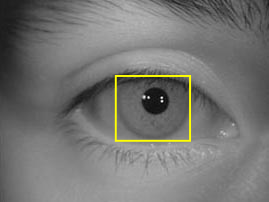}
		\caption{}
		\label{fig:iris_detection}
	\end{subfigure}
	\begin{subfigure}[b]{0.18\textwidth}
		\includegraphics[width=\textwidth]{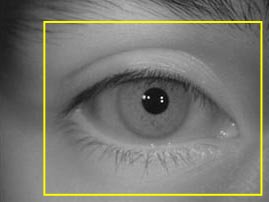}
		\caption{}
		\label{fig:eye_detection}
	\end{subfigure}
	\begin{subfigure}[b]{0.18\textwidth}
		\includegraphics[width=\textwidth]{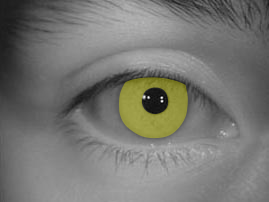}
		\caption{}
		\label{fig:iris_segmentation}
	\end{subfigure}
	\begin{subfigure}[b]{0.18\textwidth}
		\includegraphics[width=\textwidth]{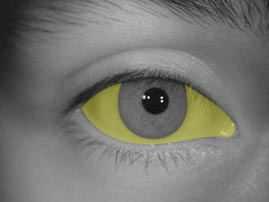}
		\caption{}
		\label{fig:sclera_segmentation}
	\end{subfigure}
	\caption{Detection and segmentation samples: (a) circular iris detection~\cite{daugman:2007}; (b) rectangular iris detection~\cite{severo2018benchmark}; (c) eye detection~\cite{lucio2019simultaneous}; (d) iris segmentation~\cite{bezerra2018robust}; (e) sclera segmentation~\cite{lucio2018fully};
	\label{fig:samples_of_delimitation}}
\end{figure*}

As stated previously, the ocular region components can be employed as input to biometrics, once these components present a high level of differentiation between users~\cite{wildes1997iris,jain_Ross_Prabhakar_2004, das2016ssrbc, das2017sserbc, delna2016sclera}. 
However, the images need to be submitted to a preprocessing stage to extract the \gls{roi}.
The preprocessing stage has great importance in a biometric system, because if the \gls{roi} extraction is erroneously performed, the system effectiveness may be injured (patterns can be removed and/or introduced into the \gls{roi})~\cite{lucio2018fully,Rattani_2017_Survey_ocular_biometrics}. 

Taking into account the importance of the preprocessing stage in the ocular region based biometrics, many different approaches were proposed to solve the \gls{roi} extraction problem. The most common approaches employed to solve the \gls{roi} extraction problem in the ocular region can be seen in ground truth samples present in Fig.~\ref{fig:samples_of_delimitation}. 
Fig. \ref{fig:circular_iris_detection} presents one of the first attempts of iris segmentation, where the \gls{roi} was delimited by the internal and external contours of the iris \cite{Liu_Bowyer_Kevin_Flynn_2005_WAIAT,daugman:2007}.
Fig.~\ref{fig:iris_detection} and Fig.~\ref{fig:eye_detection} present the delimitation approaches employed to detect the iris and eye  respectively~\cite{severo2018benchmark,lucio2019simultaneous}.
Fig.~\ref{fig:iris_segmentation} and Fig.~\ref{fig:sclera_segmentation} present the delimitation approaches employed to segment the iris and sclera respectively~\cite{bezerra2018robust,lucio2018fully}.

The most commonly employed preprocessing steps are:
contour detection~\cite{Liu_Bowyer_Kevin_Flynn_2005_WAIAT},  Hough transform~\cite{Proenca_Alexandre_2006_VISP}, active contours~\cite{Ouabida_2017_iris,Shah_Ross_2009_TIFS}, integro-differential equation~\cite{Tan2010}, \gls*{mrs}~\cite{Podder_Khan_Khan_Rahman_Ahmed_2015_ICCCI}, \glspl*{mtm}~\cite{Haindl_Krupicka_2015}, \glspl*{cnn}~\cite{lucio2018fully, bezerra2018robust, Jalilian_2017_domain_adaptation,Liu_Zhang_2016_cnn_iris,lucio2019simultaneous, zanlorensi2020attnormalization}.

\begin{figure}[!h]
	\centering

	\begin{subfigure}[b]{0.8    \linewidth}
		\centering
		\begin{subfigure}[b]{0.45\textwidth}
			\includegraphics[width=\linewidth]{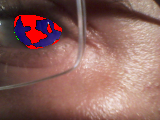}
			\caption{}
		    \label{fig:iris_seg_poor}
		\end{subfigure}%
		\hspace{4mm} 
		\begin{subfigure}[b]{0.45\textwidth}
			\includegraphics[width=\linewidth]{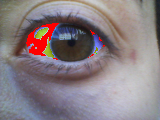}
			\caption{}
		    \label{fig:sclera_seg_poor}
		\end{subfigure}%
		
	\end{subfigure}%

	\caption{Samples of bad iris and sclera segmentation in experiments using \glspl{fcn}. The green regions in images show wrong segmented regions , while the red regions shows not segmented regions. }
	\label{fig:bad_detection_segmentation}
\end{figure}

Besides the high number of component-based preprocessing segmentation approaches, none of them works correctly in specific scenarios, such as in non-controlled environments. Performing an analysis of why the segmentation of the presented images in Fig.~\ref{fig:bad_detection_segmentation} is so bad, it was found that this issue occurs because the \gls{fcn} approach is not able to deal very well with blur and specular highlights present in the images.



The use of ~\glspl{cnn} to extract the \gls*{roi}s, taking into account context information is a promising approach, due to: (i) the available segmentation approaches do not work correctly in some scenarios; 
(ii) the context in which the \gls{roi} extraction is performed is not used by currently available approaches;
(iii) the capabilities of decision power presented by the deep learning approaches.
We can understand how the objects in a scene are related using contextual information. 

We evaluate three different types of context in order to understand the objects relationship, i.e.,
semantic (\textit{evaluates the likelihood  of an object to be found in some scenes but not in others}), spatial (\textit{evaluates the placement among objects in a scene}) and scale (\textit{evaluates the size ratio among the different classes of objects in a scene}).

 Thus, the main question addressed in this work is ``Can we improve the \gls{roi} extractor in ocular region components taking into account the context information present in an image?''

Aiming to answer this question, we propose the \gls{ercnet} as a new segmentation approach to simultaneous segment the sclera and iris using contextual information. The proposed approach combines the \gls{encnet} with the main contribution of this work, the \gls{pcloss}. The \acrshort{pcloss} consists of punishing the segmentation $Losses$ of a network by using a percentage difference value between the ground truth and the segmented masks. 
We obtain the percentage difference by taking into account Biederman’s semantic relationship concepts, in which we use three contexts (semantic, spatial, and scale) to evaluate the relationships of the objects in an image.

\begin{figure}[!h]
	\centering

	\begin{subfigure}[b]{\linewidth}
		\centering
		\begin{subfigure}[b]{0.3\textwidth}
			\includegraphics[width=\linewidth]{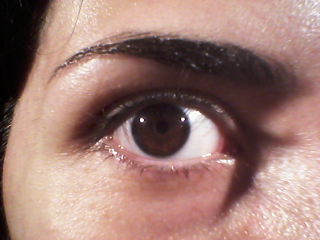}
			\caption{Miche image}
		    \label{fig:real_image}
		\end{subfigure}%
		\hspace{2mm} 
		\begin{subfigure}[b]{0.3\textwidth}
			\includegraphics[width=\linewidth]{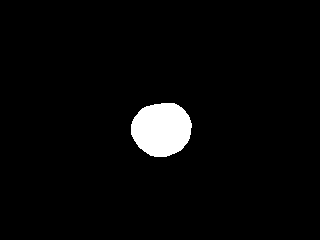}
			\caption{Iris mask}
		    \label{fig:iris_mask}
		\end{subfigure}%
		\hspace{2mm} 
		\begin{subfigure}[b]{0.3\textwidth}
			\includegraphics[width=\linewidth]{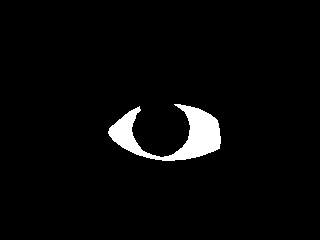}
			\caption{Sclera mask}
		    \label{fig:sclera_mask}
		\end{subfigure}%
		
	\end{subfigure}%

	\caption{Sample of iris and sclera masks for an image}
	\label{fig:annotation_sample}
\end{figure}

The main contributions of this paper are: 
(i) the \gls{pcloss} that consists into punishing the segmentation $Loss$ of the network by using a percentage value obtained, taking into account the Biederman's\cite{biederman1972perceiving} semantic relationships concepts, 
(ii) $3,191$ new manually labeled masks of the \miche dataset (sample of iris and sclera image can be seen in Fig.~\ref{fig:annotation_sample}), that is one of the most known datasets in the biometrics scenario, (iii) the novel \gls{ercnet} architecture that combines the \textit{Context Encoding Module} proposed in \cite{zhang2018context} and the \gls{pcloss} punishing approach presented in this work.

The remainder of this work is organized as follows: 
we review related works in Section~\ref{sec:related_works}. 
In Section~\ref{sec:methodology}, there is a description of the proposed methodology to perform iris and sclera segmentation using the contextual information of the image. 
Section~\ref{sec:experiments} presents the datasets, evaluation protocol and baselines used in the experiments. 
We report and discuss the results in Section~\ref{sec:results_and_discussion}.
We state our conclusions in Section~\ref{sec:conclusion}.


\section{Related Works}
\label{sec:related_works}

In this section, we review several approaches 
for ocular region components extraction.
More specifically, we discuss works related to segmentation in Section~\ref{sec:segmentation}, and in Section~\ref{sec:contextual}, we present works in which the contextual relationship among the elements present in the image are explored.

\subsection{Segmentation}
\label{sec:segmentation}

As previously stated, segmentation has great importance in the biometric systems. 
This importance comes from the fact that an incorrect extraction of the~\gls{roi} can affect the effectiveness of the whole system.

\glsreset{cnn}

Inspired by the excellent results obtained by using \gls{cnn} approach in computer vision domain, many authors have proposed \gls{cnn} based approaches to extract the ocular region components~\cite{Liu_Zhang_2016_cnn_iris,bezerra2018robust,osorio_2018,rot_2018,zhang2014new}.

Liu et al.~\cite{Liu_Zhang_2016_cnn_iris} was one of the first authors to propose the use of \gls{cnn} to iris segmentation.
They proposed two new approaches called \glspl*{hcnn} and \glspl*{mfcn}. Performing a dense prediction of the pixels using sliding windows, merging shallow and deep layers, they achieved $0.90\%$ and $0.59\%$ \glspl{eer} on the UBIRIS V2 and CASIA-IrisV4-Distance~\cite{CASIA2010} databases.

Bezerra et al.~\cite{bezerra2018robust} presented a \gls{fcn}-based approach, and achieved state-of-the-art results on the employed databases, i.e., BioSec~\cite{Fierrez2007}, CASIA-IrisV3-Interval, CASIA-IrisV4-Thousand, \gls{iiitdcli}\cite{dhamecha2014recognizing}, NICE.I~\cite{Proenca2012}, CROSS-EYED~\cite{Sequeira2016} and MICHE-I~\cite{DeMarsico2015}, obtaining \glspl{eer} of $0.58\%$, $0.55\%$, $1.25\%$, $00.72\%$, $2.67\%$, $1.12\%$, and $1.90\%$, respectively.

Osorio-Roig et al.~\cite{osorio_2018}, similarly to Bezerra et al.~\cite{bezerra2018robust},  have also presented a \gls{fcn} approach to iris segmentation. In the proposed approach , the authors employed a multi-class training, however, only the iris segmentation was evaluated. To evaluate the proposed approach, the \nicei competition protocol was employed in two well-know iris datasets, \mobbio~\cite{mobbio_2014} and \nicei, obtaining \glspl{eer} of $2.46\%$ and $1.13\%$, respectively.

Zhao  and  Kumar~\cite{zhao_2019} proposed the \uninet, a new iris segmentation approach, and just like Osorio-Roig et al.~\cite{osorio_2018} used the \nicei competition protocol to evaluate the obtained results. The datasets employed to evaluate the \uninet are: \ndiris, \distance and \iitd, obtaining \glspl{eer} of $1.68\%$, $0.67\%$, and $5.34\%$, respectively.

Despite the consolidation of the iris as a biometric resource, other elements that compose the ocular region have started to arouse the interest of the researchers. 
Besides the iris, the sclera was one of the first ocular region components to attract the attention of the researchers, once the biometric systems based on such trait achieved interesting results~\cite{delna2016sclera}.
Some works were proposed~\cite{lucio2018fully,rot_2018, wang_2019} taking into account the good results on sclera based biometrics and the need for a proper delimitation of the \gls{roi}, as well as in the iris-based biometrics.

Lucio et al.~\cite{lucio2018fully} proposed two new approaches to sclera segmentation. 
The first one is based on the \glspl*{gan}~\cite{goodfellow2014generative}, and the second approach is based on \glsreset{fcn}\glspl*{fcn}~\cite{long2015fully}. 
The best results were obtained using the \acrshort{fcn} approach achieving F-Score values of $87.48\%$, $88.32\%$, $88.12\%$, $87.80\%$ and $87.94\%$  on the UBIRIS V2, MICHE-I, MICHE-GS4, MICHE-IP5, and MICHE-GT2 databases, respectively. 

Rot el al.~\cite{rot_2018} proposed a multi-class segmentation approach based in the \segnet to segment simultaneously the sclera and another components from the periocular region (iris, pupil, periocular, eyelashes and canthus). The authors employed the \gls{masd} in the evaluation of the proposed approach, obtaining F-Score vales of $91.00\%$, $91.00\%$, $85.00\%$, $90.00\%$, $63.00\%$ and $53.00\%$ on the iris, sclera, pupil, periocular, eyelashes and canthus regions, respectively.

Wang et al~\cite{wang_2019} presenteda new sclera segmentation approach based in the \unet, and achieved F-Score vales of $91.43\%$, $89.54\%$, $90.45\%$, $89.28\%$ and $89.34\%$ on the UBIRIS V2, MICHE-I, MICHE-GS4, MICHE-IP5, and MICHE-GT2 databases, respectively.

Taking into account the metrics presented to evaluate the segmentation results on ocular region components, the one that best quantifies how well a \acrshort{roi} is segmented is the \glsreset{iou}\gls{iou} metric. 
The use of \gls{iou} as a evaluation metric is based on the fact that using this it is possible to compare how close the segmented masks are to the ground-truth.

\subsection{Region of Interest Extraction Using Contextual Relationship}
\label{sec:contextual}

In addition to the commonly employed approaches to extract the ocular region components presented in the previous sections, there is a set of techniques that aims to detect/segment elements in an image based on the contextual information in which it is located. 
The contextual information can be defined as any information that is not directly produced by the appearance of an object, such as, nearby data, image tags, annotations or the presence of other objects~\cite{kumar_2005,Hoiem2008,tu_2010}.

A traditional object categorization approach is made using visual attributes such as color, edges responses, texture, and shapes cues. 
However, according to Biederman~\cite{biederman1972perceiving}, using only these feature sets, understanding a scene composition is not possible. 
To detect an object based on context , it is necessary to understand the five different classes of relationships between an object and its surroundings: \textit{interposition}, \textit{support}, \textit{probability}, \textit{position}, and \textit{size}. 
\textit{Interposition} and \textit{support} refer to the physical space. \textit{Probability}, \textit{position}, and \textit{size} are defined as semantic relations once they require access to the referential meaning of the object.
Semantic relations include information about specific interactions among objects in the scene, and they are often used as \textit{contextual features}.

Many authors have explored the semantic relationships proposed in \cite{biederman1972perceiving} aiming to improve the results on recognition task of objects~\cite{Torralba2003, Fink_2003,carbonetto2004statistical,He_2004, Rabinovich_2007, Galleguillos_2008}.
These relationships can be grouped into three categories: semantic context (\textit{probability}), spatial context (\textit{position}), and scale context (\textit{size}). 

Semantic context corresponds to the likelihood of an object to be found in some scenes but not in others. 
Hence, it is defined in terms of the co-occurrence of one object with others and its occurrence in scenes. 

Torralba~\cite{Torralba2003}~presented one of the first methods using statistical approaches to detect objects. 
The proposed technique explores and generalizes the semantic context in a real world scenario, by using the correlation among the statistics of low-level features across the entire scene and the objects it contains.
However, the authors do not report either the obtained results or the database employed in this work. 

In the same direction of the work proposed by Torralba~\cite{Torralba2003}, other authors have employed statistical methods to detect objects based on semantic context~\cite{Wolf_2006, Rabinovich_2007, Verbeek_2007, Galleguillos_2008}.

Wolf and Bileschi~\cite{Wolf_2006} proposed an approach to extract the semantic context using context features and the \glsreset{svm}\gls{svm} classifier.
The presented approach achieved an accuracy of $83.00\%$ (estimated value obtained from a Receiver Operating Characteristics (ROC) curve on that work) on their database named StreetScene. 
To obtain the context features, the authors proposed a two-stage process. 
In the first stage, the image is processed to calculate the low level and semantic information.
In the second stage, the context feature is calculated at each point by collecting samples of the previously computed features at predefined relative positions.

Rabinovich et al.~\cite{Rabinovich_2007} employed the \glsreset{crf}\gls*{crf} to detect the objects in a scene maximizing the labels agreement according to the contextual relevance.
Firstly, a fully connected graph among the segmented labels was used.
And then, the \acrshort*{crf} was trained in more straightforward problems defined on a relatively small number of segments. 
The proposed approach showed to be competitive with the state-of-the-art achieving $74.20\%$ and $68.40\%$ accuracies on the Pascal VOC 2007~\cite{pascal-voc-2007} and \gls{msrc}~\cite{msrc} databases, respectively.




Verbeek and Triggs~\cite{Verbeek_2007} proposed a segmentation approach that achieves  $84.90\%$, $87.40\%$ and $74.60\%$ of accuracy on the \acrshort{msrc}, Sowerby~\cite{He_2004} and Corel\footnote{https://sites.google.com/site/dctresearch/Home/content-based-image-retrieval} databases respectively. 
These results were obtained using \acrshort{crf} together with unlabeled nodes.
This way the unknown labels are marginalized and the log-likehood of the known labels can be maximized by gradient descendent.

Galleguillos et al.~\cite{Galleguillos_2008} introduced a new approach to object categorization based on two types of context (co-occurrence and relative location) using local appearance features, and achieved $36.70\%$ and $68.47\%$ of accuracy, on the Pascal Voc 2007 and \acrshort{msrc} databases respectively. 
The presented method used \glspl*{crf} to maximize object label attribution to both semantic and spatial features. The results of that work showed that combining co-occurrence and spatial context improves the accuracy of the system when compared to the results where only co-occurrence data are used in training.

\begin{figure}[!htb]
	\centering
	\includegraphics[width=0.6\linewidth]{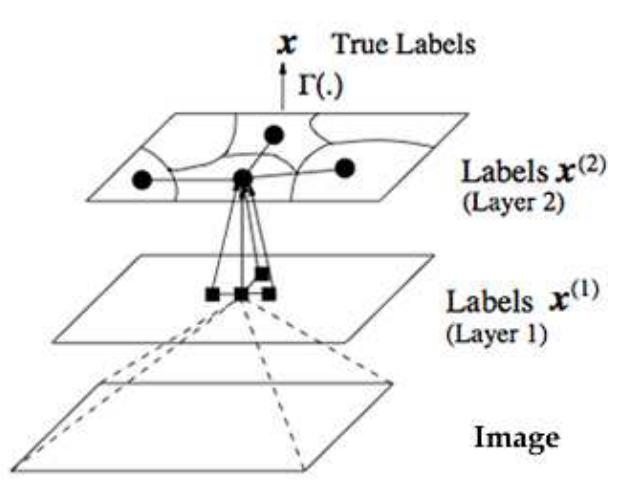}
	\caption{Sample of the architecture proposed by \cite{kumar_2005}.}
	\label{fig:kumar_layers}
\end{figure}

Since the semantic context has been well explored in the literature, Kumar and Hebert ~\cite{kumar_2005} proposed a new segmentation approach that explored the spatial context in the images. The proposed approach explore the pairwise relationship in images using a two-layer hierarchical formulation to exploit different levels of contextual information in images. 
In the first layer, the system it encodes the region's interaction, and in the second, the object's interactions are mapped, as shown in Figure \ref{fig:kumar_layers}.
The experiments were performed on four databases: Beach Database~\cite{kumar2003observation}, Sowerby Database~\cite{He_2004}, Building/Road/Car dataset~\cite{Torralba:2004} and Monitor/Keyboard/Mouse Dataset~\cite{Torralba:2004}, and the respective accuracies for these are  $74.00\%$, $89.30\%$, $84.36\%$ and $90.00\%$ respectively (result obtained analyzing a \gls{roc} curve).

As we can see, the semantic context is implicitly present in the spatial context, since the information of object co-occurrences come from identifying objects from the spatial relations in the scene. The same happens to scale context, once a contextual relationship establishes that objects have a limited set of size relationships with other objects in the scene. Therefore, the use of spatial and scale context involves using all forms of contextual information in the scene.

Over the years, many approaches have been proposed to extract a region of interest \gls{roi} using contextual information. 
But, only after $2018$ the first works that integrate the Biederman's~\cite{biederman1972perceiving} contextual classes with \glspl{cnn} were proposed.
Leng et al.~\cite{Leng_2018} introduced the \glsreset{cln}\gls{cln}, which aims to capture the pairwise relationships between objects and the global context of each one. 
By using the proposed method, the authors achieved $82.10\%$, $80.70\%$, and $38.40\%$ of \glsreset{map}\gls{map} on the Pascal VOC 2007, Pascal VOC 2012, and Coco databases, respectively. 
The \gls{cln} is composed of two subnetworks, a \gls{mlp} which captures the pairwise relationships and a \acrshort{cnn} that learns the global context of the image.

Leng et al.~\cite{Leng_2019} also proposed a context-aware U-Net, which focuses on capturing valuable contexts and improving the segmentation performance on the ISBI Challenge Database. 
For such goal, a lightweight context transfer module was developed to learn the rich context features present in the image. The results obtained are similar to those presented by the state-of-the-art on the ISBI Challenge.
To evaluate the results, the authors proposed three metrics: Warping Error, Rand Error, and Pixel Error, their respective achieved values are $1.212e^{-4}\%$, $2.12e^{-2}\%\%$, and $3.4600e^{-2}\%\%$.

\subsection{Final Remarks}
\label{sec:final_remarks}

Since, to the best of our knowledge,  the specific works present in the literature for the ocular region components detection do not address the use of contextual information, and the works that made use of this approach in the scene understanding scenario present promising results, it is opportune to develop a specific approach for the extraction of the \gls{roi} in the ocular region taking into account the contextual information.

\section{Methodology}
\label{sec:methodology}

\begin{figure*}[!htb]
	\begin{center}
		\includegraphics[width=1\linewidth]{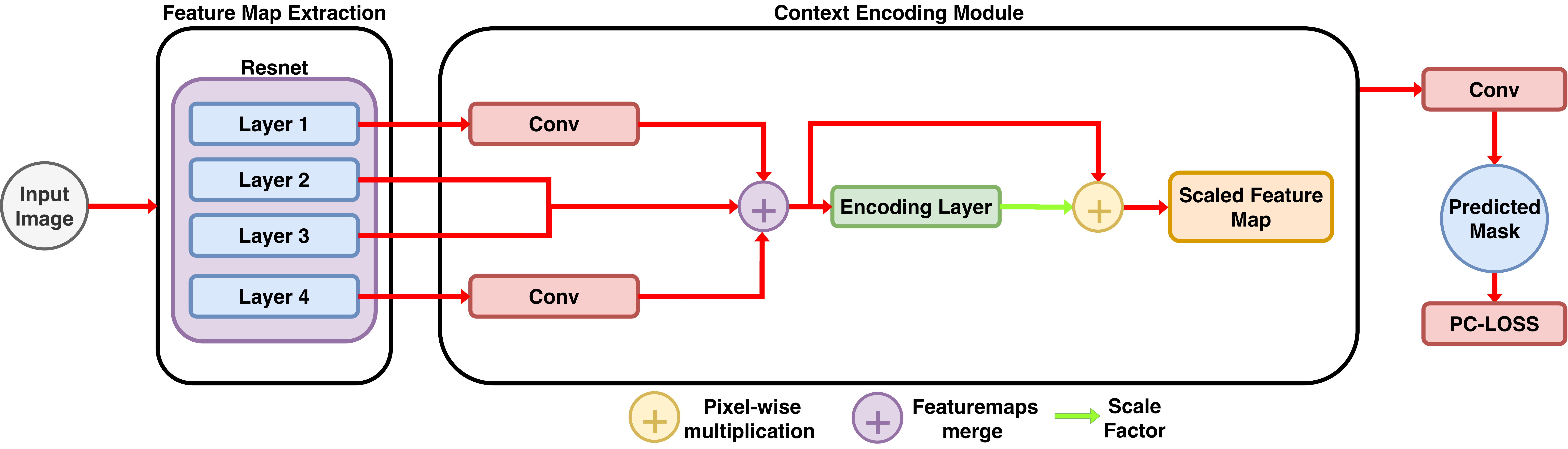}
		\caption{General overview of the proposed approach.}
		\label{fig:general_overview_of_the_proposed_approach}
	\end{center}
\end{figure*}

This section presents the proposed architecture addressing the ocular region components (iris and sclera) segmentation, taking into account the contextual information of the region.

To solve the aforementioned segmentation problem, we propose the \glsreset{ercnet}\gls{ercnet}, a new architecture that combines the \glsreset{encnet}\gls{encnet} presented in \cite{zhang2018context} with the main contribution of this work: the \glsreset{pcloss}\gls*{pcloss}. \gls*{pcloss} was inspired in the Biederman’s semantic relationship concepts in which the relationships of objects in an image are evaluated using semantic, scale and spatial context.

In the \gls{encnet}, the extraction of feature maps is performed by using a \resnet based backbone, which was chosen because it does not suffer the effects of the vanishing gradient, thus it does not compromise the network overall performance.
Once the feature maps are extracted using the \resnet backbone, they were used to feed the \textit{Context Encoding Module}.

The \textit{Context Encoding Module} is composed of an \textit{Encoding Layer}~\cite{zhang2017deep} which captures the encoded semantics, a fully connected layer, and a sigmoid as the activation function, which output scaling factors  $\gamma=\delta(W_e)$, where $W$ denotes the layer weights and $\delta$ is the sigmoid function. 
Then, the module output is given by $Y=X\otimes\gamma$, where $\otimes$ denotes a channel-wise multiplication between the input feature maps $X$ and scaling factors $\gamma$. 
The channel-wise multiplication is employed to emphasize or de-emphasize class-dependent feature maps.

Once the feature maps are processed by the \acrshort{encnet}, we employ the \glsreset{pcloss}\gls{pcloss}.
The \gls{pcloss} consists of punishing the segmentation $Loss$ of the network by using a percentage value obtained, taking into account the Biederman's\cite{biederman1972perceiving} relationships, i.e., 
\begin{equation}
\text{\gls{pcloss}} = \frac{\lambda + \rho}{2},
\end{equation}
\noindent where the parameter $\lambda$ is the \textit{Scale Context Coefficient} (a value that estimates the average difference in terms of \textit{Jaccard Distance} from a class to all other classes on the addressed problem) of an image and it is defined as:

\begin{equation}
\lambda = \frac{1}{N}\sum_{i=1}^{N}\frac{1}{N-1}\underset{j \ne i}{\sum_{j=1}^{N}}\sqrt[2]{(\theta_i-\theta_j)^2},
\end{equation}

\noindent where $N$ is the number of classes present in the evaluated problem,
and the $\theta$  term is the \textit{Jaccard Distance} obtained by subtracting the \textit{Jaccard similarity coefficient} $J = \frac{gt_i \cap prd_i}{gt_i \cup prd_i}$ from $1$, where $gt_i$ and $prd_i$ stand for the ground truth and predicted masks, respectively, i.e.,
\begin{equation}
\theta_i = 1-J.
\end{equation}
\noindent 
The $\rho$ parameter is the ratio between the \textit{Spatial Context Coefficient} $\delta$ from the ground truth and the predicted segmentation,
\begin{equation}
\rho = \frac{\delta(prd)}{\delta(gt)}.
\end{equation}
\noindent where
\begin{equation}
\delta = \frac{1}{N}\sum_{i=1}^{N}\frac{1}{N-1}\underset{i \ne j}{\sum_{j=1}^{N}}\sqrt[2]{(C_i-C_j)^2},
\end{equation}
\noindent 
$C$ the center of mass from an evaluated object.
Considering that an object consists of $n$ distinct points $x_1 ... x_n$, then the centroid (center of mass) is given by,
\begin{equation}
C = \frac{1}{n}\sum_{i=1}^{n}x_i.
\end{equation}
The \gls{ercnet} implementation was made in the PyTorch framework, and to perform the training we adopted a learning rate of $10^{-4}$ combined with a momentum of $0.9$ during $1000$ epochs.


\section{Experiments}
\label{sec:experiments}

In this section, we present the database, the evaluation protocol and the state-of-the-art segmentation approaches employed as baselines  used in this work.
The experiments were carried out on the \miche database, which is described in Section~\ref{sub_sec:databases}.
Note that we evaluate the networks on each \miche subset separately.
All experiments were performed on a computer with an AMD\textsuperscript{\textregistered} Ryzen™ $7$ $2700$X $4.40$GHz CPU, $32$ GB of RAM and two NVIDIA Titan V GPUs.

\begin{figure*}[!ht]
	\captionsetup[subfigure]{justification=centering}

	\begin{subfigure}{1\linewidth}
		\centering
		\begin{subfigure}[t]{.105\linewidth}
			\includegraphics[width=1\linewidth]{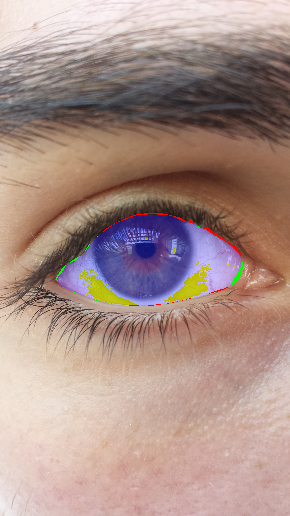}
			\caption*{EncNet \\ \resnet-$50$}
			
		\end{subfigure}
		\begin{subfigure}[t]{.105\textwidth}
			\includegraphics[width=1\linewidth]{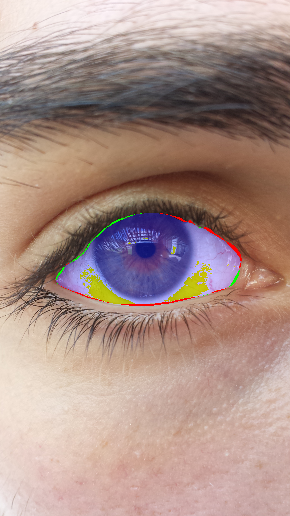}  
			\caption*{EncNet \\ \resnet-$101$}
		\end{subfigure}
		\begin{subfigure}[t]{.105\textwidth}
			\includegraphics[width=1\linewidth]{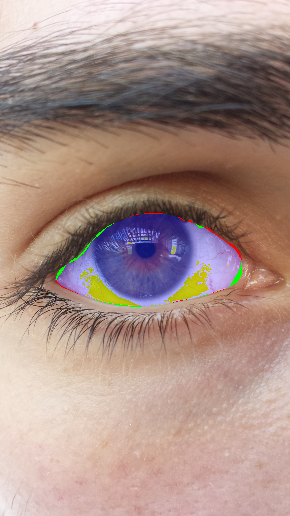}
			\caption*{EncNet \\ \resnet-$152$}  
		\end{subfigure}
		\begin{subfigure}[t]{.105\textwidth}
			\includegraphics[width=1\linewidth]{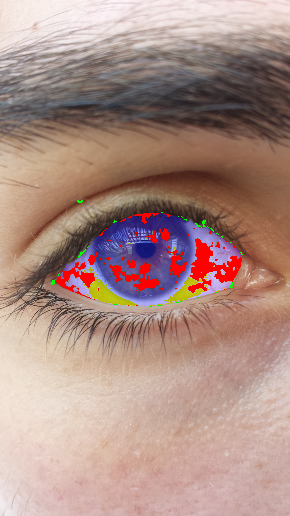}
			\caption*{\gls{fcn}}  
		\end{subfigure}
		\begin{subfigure}[t]{.105\textwidth}
			\includegraphics[width=1\linewidth]{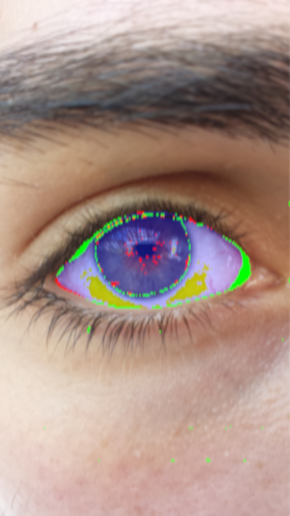}
			\caption*{\gls{gan}}  
		\end{subfigure}
		\begin{subfigure}[t]{.105\textwidth}
			\includegraphics[width=1\linewidth]{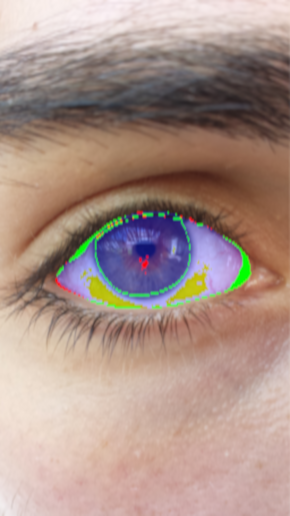}
			\caption*{Processed \gls{gan}} 
		\end{subfigure}
		\begin{subfigure}[t]{.105\textwidth}
			\includegraphics[width=1\linewidth]{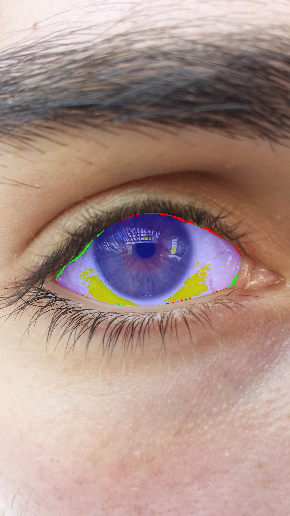}
			\caption*{\gls{ercnet} \\ \resnet-$50$}  
		\end{subfigure}
		\begin{subfigure}[t]{.105\textwidth}
			\includegraphics[width=1\linewidth]{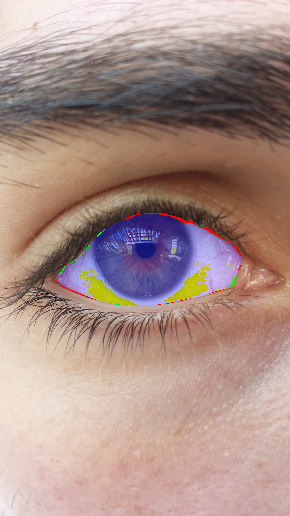}
			\caption*{\gls{ercnet} \\ \resnet-$101$}  
		\end{subfigure}
		\begin{subfigure}[t]{.105\textwidth}
			\includegraphics[width=1\linewidth]{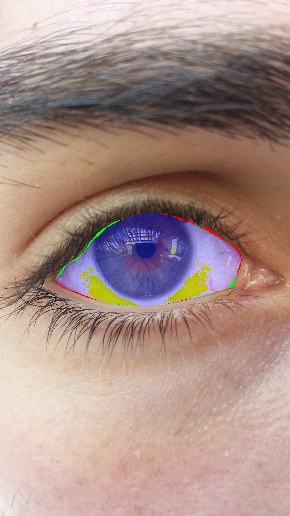}
			\caption*{\gls{ercnet} \\ \resnet-$152$}  
		\end{subfigure}
		\caption{All scenario segmentation qualitative results}
	\end{subfigure}
	
	\begin{subfigure}{1\linewidth}
		\centering
		\begin{subfigure}[t]{.105\linewidth}
			\includegraphics[width=1\linewidth]{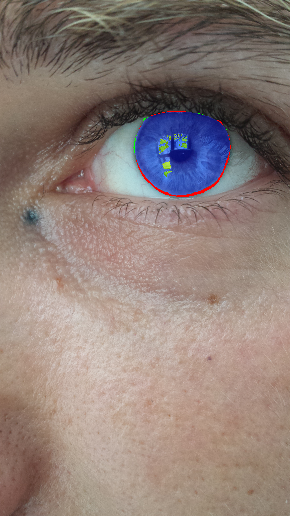}
			\caption*{EncNet \\ \resnet-$50$}
			
		\end{subfigure}
		\begin{subfigure}[t]{.105\textwidth}
			\includegraphics[width=1\linewidth]{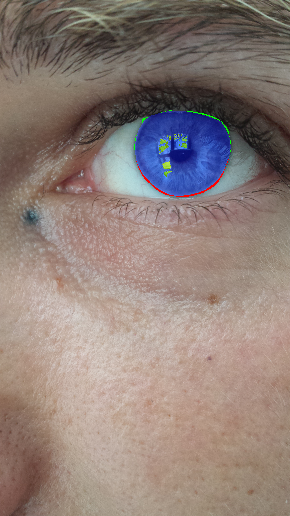}
			\caption*{EncNet \\ \resnet-$101$}
		\end{subfigure}
		\begin{subfigure}[t]{.105\textwidth}
			\includegraphics[width=1\linewidth]{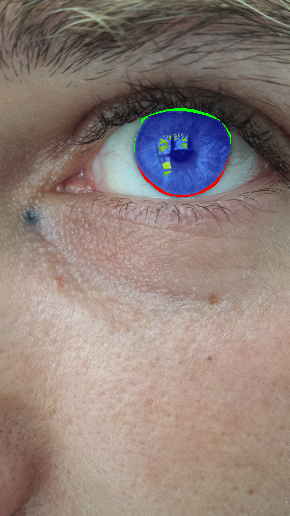}
			\caption*{EncNet \\ \resnet-$152$}  
		\end{subfigure}
		\begin{subfigure}[t]{.105\textwidth}
			\includegraphics[width=1\linewidth]{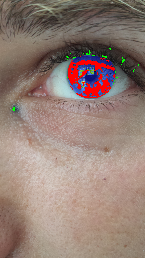}
			\caption*{\gls{fcn}}  
		\end{subfigure}
		\begin{subfigure}[t]{.105\textwidth}
			\includegraphics[width=1\linewidth]{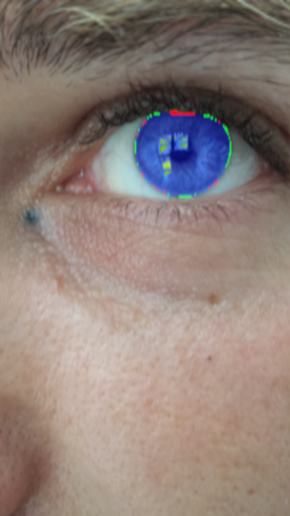}
			\caption*{\gls{gan}}  
		\end{subfigure}
		\begin{subfigure}[t]{.105\textwidth}
			\includegraphics[width=1\linewidth]{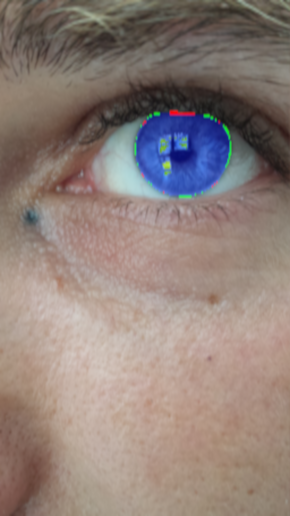}
			\caption*{Processed \gls{gan}} 
		\end{subfigure}
		\begin{subfigure}[t]{.105\textwidth}
			\includegraphics[width=1\linewidth]{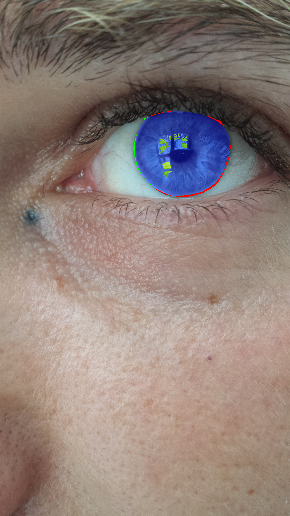}
			\caption*{\gls{ercnet} \\ \resnet-$50$}  
		\end{subfigure}
		\begin{subfigure}[t]{.105\textwidth}
			\includegraphics[width=1\linewidth]{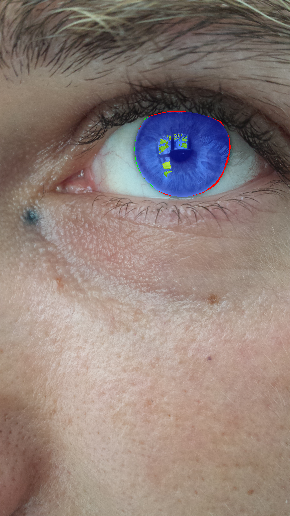}
			\caption*{\gls{ercnet} \\ \resnet-$101$}  
		\end{subfigure}
		\begin{subfigure}[t]{.105\textwidth}
			\includegraphics[width=1\linewidth]{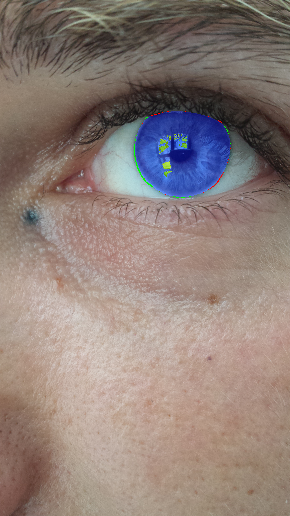}
			\caption*{\gls{ercnet} \\ \resnet-$152$}  
		\end{subfigure}
		\caption{Iris segmentation qualitative results}
	\end{subfigure}
	
	\begin{subfigure}{1\linewidth}
		\centering
		\begin{subfigure}[t]{.105\linewidth}
			\includegraphics[width=1\linewidth]{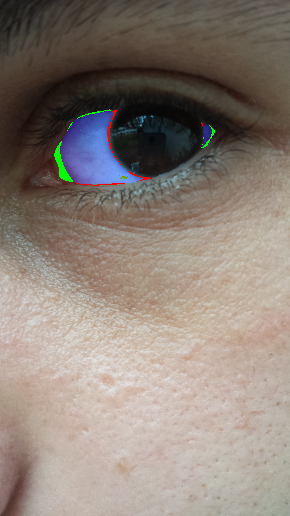}
			\caption*{EncNet \\ \resnet-$50$}
			
		\end{subfigure}
		\begin{subfigure}[t]{.105\textwidth}
			\includegraphics[width=1\linewidth]{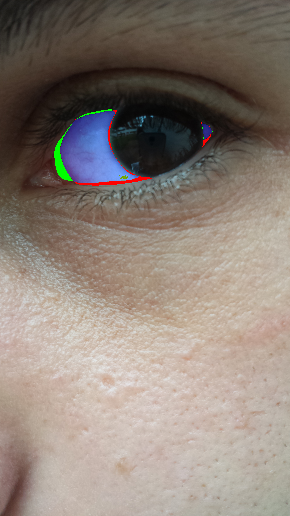}
			\caption*{EncNet \\ \resnet-$101$}
		\end{subfigure}
		\begin{subfigure}[t]{.105\textwidth}
			\includegraphics[width=1\linewidth]{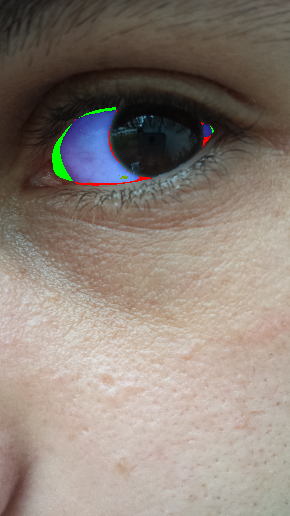}
			\caption*{EncNet \\ \resnet-$152$}  
		\end{subfigure}
		\begin{subfigure}[t]{.105\textwidth}
			\includegraphics[width=1\linewidth]{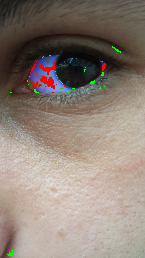}
			\caption*{\gls{fcn}}  
		\end{subfigure}
		\begin{subfigure}[t]{.105\textwidth}
			\includegraphics[width=1\linewidth]{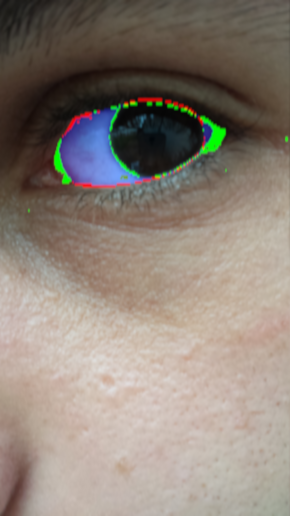}
			\caption*{\gls{gan}}  
		\end{subfigure}
		\begin{subfigure}[t]{.105\textwidth}
			\includegraphics[width=1\linewidth]{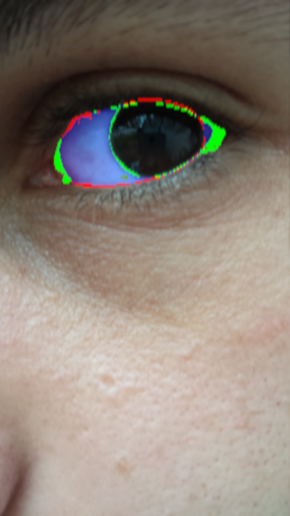}
			\caption*{Processed \gls{gan}} 
		\end{subfigure}
		\begin{subfigure}[t]{.105\textwidth}
			\includegraphics[width=1\linewidth]{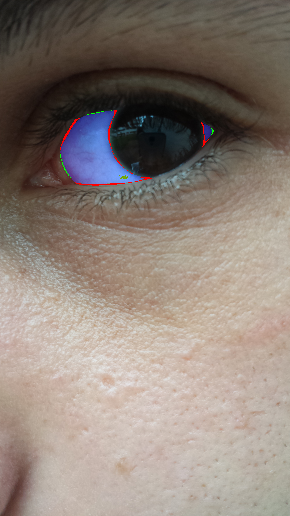}
			\caption*{\gls{ercnet} \\ \resnet-$50$}  
		\end{subfigure}
		\begin{subfigure}[t]{.105\textwidth}
			\includegraphics[width=1\linewidth]{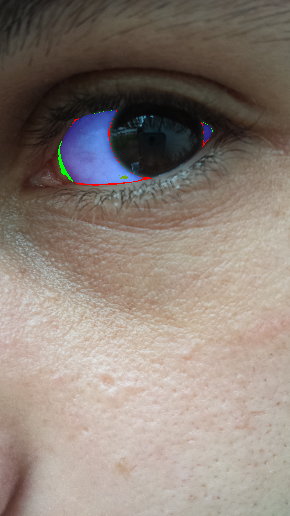}
			\caption*{\gls{ercnet} \\ \resnet-$101$}  
		\end{subfigure}
		\begin{subfigure}[t]{.105\textwidth}
			\includegraphics[width=1\linewidth]{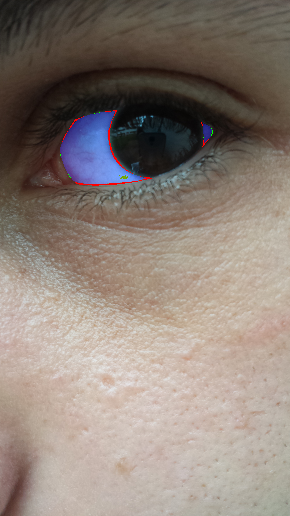}
			\caption*{\gls{ercnet} \\ \resnet-$152$}  
		\end{subfigure}
		\caption{Sclera segmentation qualitative results}
	\end{subfigure}

	\caption{Qualitative results achieved by the \gls{ercnet}, and by the proposed baselines. Green and red pixels represent \gls{fp} and \gls{fp}, repectively.}
	\label{fig:general_results_comparison}
	
\end{figure*}

\subsection{Databases}
\label{sub_sec:databases}

To evaluate the performance of our proposed approach to solve the problem of iris and sclera simultaneous segmentation, the \miche database was employed. 
The database was chosen because it is composed of $3,191$ non-processed images captured using three different devices (see Table \ref{tab:overview_datasets}). 
The devices used to capture the images of the database are: 
\begin{itemize}
	\item Samsung Galaxy S$4$ (GS$4$) smartphone: Google Android operating system, two sensors: a \gls*{cmos} sensor with $13$ Megapixel (72 dpi) and $2322\times4128$ resolution, and a \gls*{cmos} sensor with $2$ Megapixel (72\textbf{}) and $1080\times1920$ resolution;
	\item Samsung Galaxy Tab $2$ tablet: Google Android Operating System, one sensor with $0.3$ megapixel (72 dpi) and $640 \times 480$ resolution;
	\item Apple iPhone 5 smartphone: Apple iOS Operating System, two sensors: a \gls*{cmos} sensor with $8$ Megapixel (72 dpi) and $1536 \times 2048$ resolution, and a \gls*{cmos} sensor with $1.2$ Megapixel (72 dpi) and $960 \times 1280$ resolution.
\end{itemize}

\begin{table}[!htb]
	\centering
	\caption{Overview of the databases used in this work~\cite{demarsico_2015_MICHE_database}.}
	\vspace{1mm} 
	\label{tab:overview_datasets}
			\begin{tabular}{@{}cccc@{}}
				\toprule
				\textbf{Database}    & \textbf{Images} & \textbf{Subjects} & \textbf{Resolution}  \\ \midrule
				\acrshort*{gs4} & $1$,$297$ & $92$ & $2322 \times 4128$ and $1080 \times 1920$      \\
				\acrshort*{gt2} & $632$ & $92$ & $640 \times 480$      \\
				\acrshort*{ip5} & $1$,$262$ & $92$ & $1536 \times 2048$ and $960 \times 1280$      \\
				\bottomrule
			\end{tabular}
\end{table}
 
\begin{figure}[!ht]
	\centering
	\begin{subfigure}{.24\linewidth}
		\centering
		\includegraphics[width=1\linewidth]{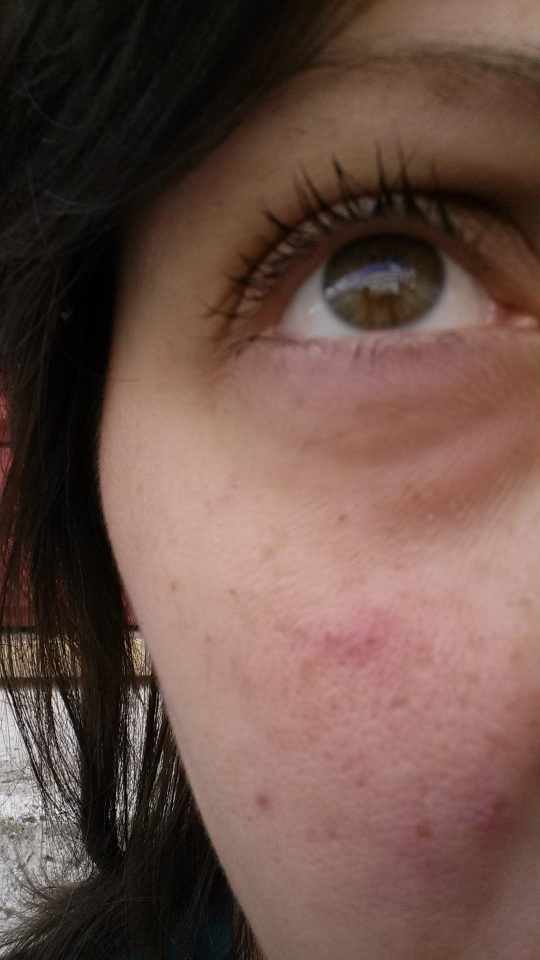}  
		\caption{}
	\end{subfigure}
	\begin{subfigure}{.24\linewidth}
		\centering
		\includegraphics[width=1\linewidth]{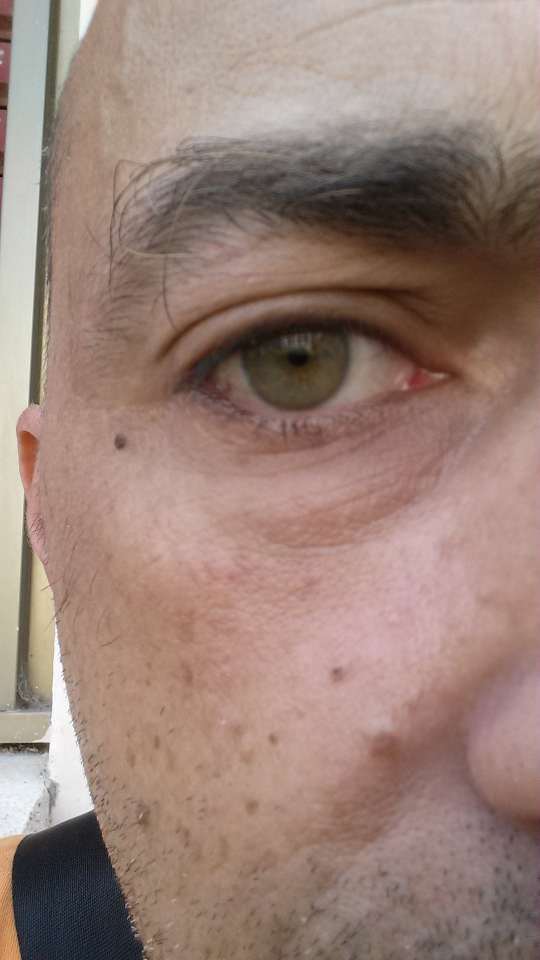}  
		\caption{}
	\end{subfigure}
	\begin{subfigure}{.24\linewidth}
		\centering
		\includegraphics[width=1\linewidth]{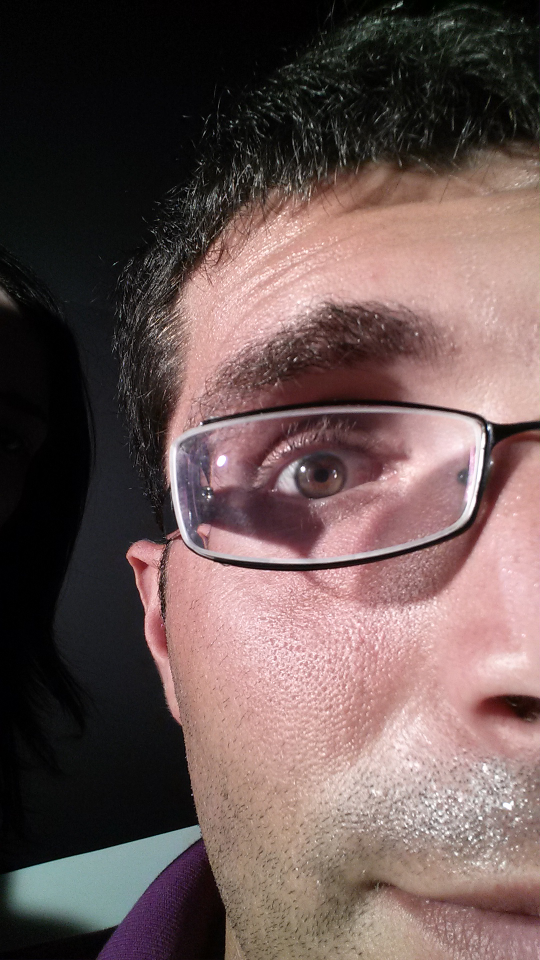}  
		\caption{}
	\end{subfigure}
	\begin{subfigure}{.24\linewidth}
		\centering
		\includegraphics[width=1\linewidth]{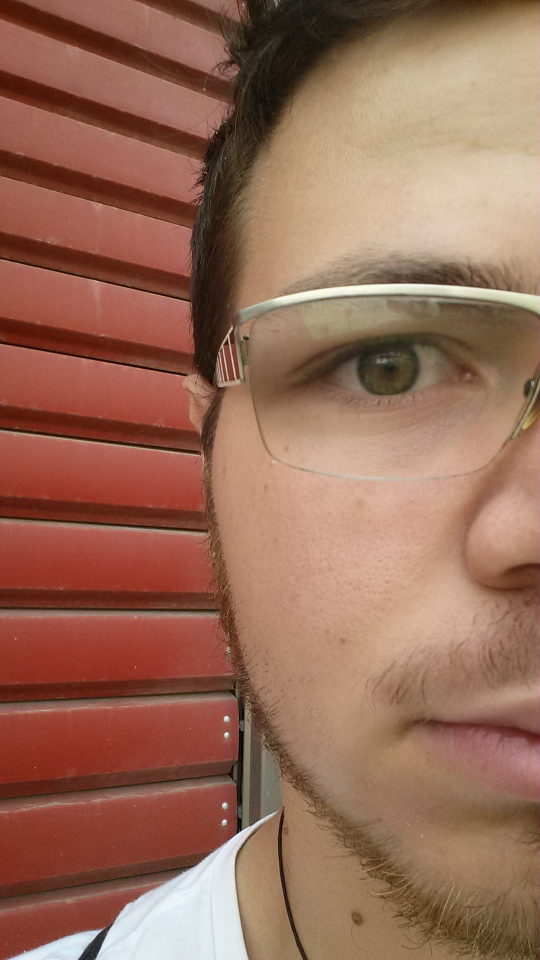}  
		\caption{}
	\end{subfigure}
	\caption{Samples of difficult images}
	\label{fig:hard_to_segment_images}
\end{figure}

The non-processed images are the ones that, in addition to the components of the ocular region, the images also have parts of the face and even other types of objects (background objects, glasses, nose, mouth) as can be seen in Fig. ~\ref{fig:hard_to_segment_images}, which makes the segmentation task even more challenging.


\subsection{Evaluation Protocol}
\label{subsec:evaluation_protocol}

The evaluation of an automatic detection approach is performed in a pixel-to-pixel comparison between the ground truth and the predicted segmentation mask.
Therefore, we use the mean $F$-score, \glsreset{iou}\gls*{iou} and \glsreset{eer}\gls*{eer} evaluation~metrics.

In order to perform a fair evaluation and comparison of the proposed approaches, we divided each database into three subsets, being $40\%$ of the images for training, $40\%$ for testing and $20\%$ for validation. 
We adopted this protocol (i.e., with a larger test set) to provide more samples for the sake of statistical significance. 
In addition, we perform the Wilcoxon signed-rank test~\cite{wilcoxon1970critical} with significance level of $\alpha=0.05$ to verify if there is a statistical difference between the detection~approaches.

\subsection{Baselines}
\label{subsec:Baseline}

We selected three baseline frameworks described (and available) in the literature to compare to our approach: \gls*{encnet}~\cite{zhang2018context}, \glsreset{fcn}\gls*{fcn}~\cite{long2015fully} and \glsreset{gan}\gls{gan}~\cite{goodfellow2014generative}. 

The \gls{fcn} and the \gls{gan} segmentation approaches were chosen as baseline methods because they presented promising results on sclera and iris segmentation tasks~\cite{bezerra2018robust,lucio2018fully}. The \gls*{encnet} approach was chosen since the \textit{Context Encoding Module} proposed in it encodes the contextual information in the image, thus outperforming the conventional segmentation approaches employed in many different research areas.

The \textbf{\gls*{fcn}} segmentation approach was proposed in \cite{long2015fully}. The network has only convolutional layers and the segmentation process can take input images of arbitrary sizes, producing correspondingly-sized output with efficient inference and learning.

\begin{figure}[!htb]
	\begin{center}
		\includegraphics[width=0.3\linewidth]{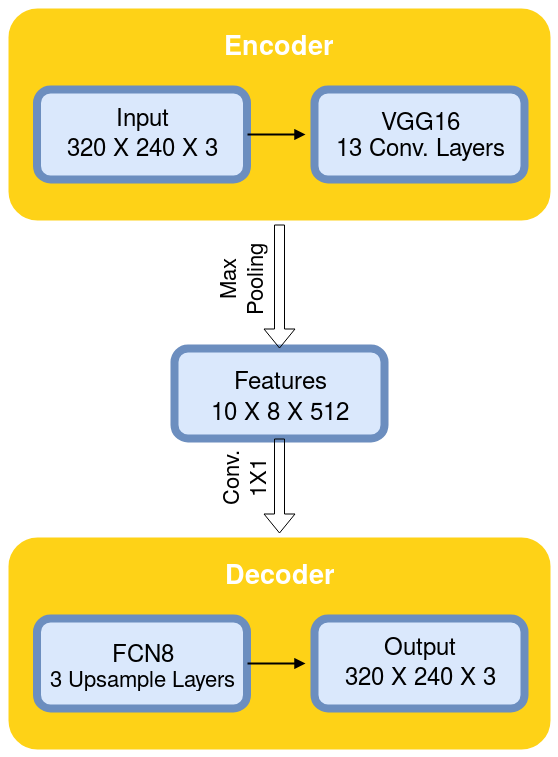}
	\end{center}
	\vspace{-2mm}
	\caption{\gls*{fcn} architecture employed in the simultaneous ocular region components segmentation.}
	\label{fig:architecture_fcn}
\end{figure}

In this work, we employ the \gls*{fcn} approach presented by Teichmann et al.~\cite{teichmann2016multinet}. As shown in Figure~\ref{fig:architecture_fcn}, features are extracted using a \gls*{cnn} without the fully connected layers (i.e., VGG-$16$ without the last $3$ layers).

Next, the extracted features pass through two $1\times1$ convolutional layers, generating an output of dimension $10\times8\times 6$. The output of these convolutional layers is processed by the FCN8 architecture proposed in~\cite{long2015fully}, which performs the up sampling combining the last three layers from the VGG-$16$. 

\begin{figure}[!h]
	\begin{center}
		\includegraphics[width=.5\linewidth]{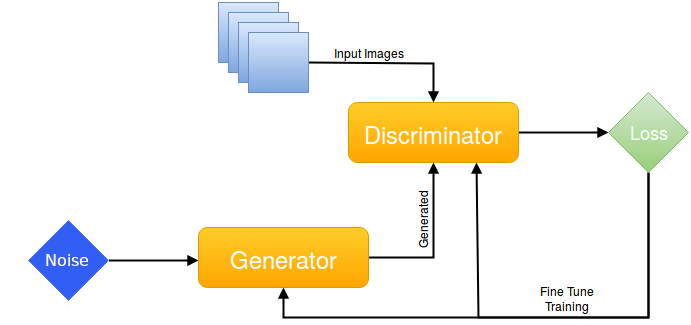}
	\end{center}
	\vspace{-2mm}
	\caption{\gls*{gan} architecture employed in the simultaneous ocular region components segmentation.}
	\label{fig:architecture_gan}
\end{figure}

\textbf{\glspl*{gan}} are a type of deep neural network composed of a generator and a discriminator networks, which pits one against the other. In a first moment the generator network receives noise as input and generates sample images, and the discriminator network evaluates how close the ground truth is to the generated images~\cite{goodfellow2014generative}.
A generic GAN architecture is shown in Fig.~\ref{fig:architecture_gan}. 

Basically, the generator network learns to produce more realistic samples throughout each iteration, and the discriminator network learns to better distinguish between real and synthetic data.

The \gls*{gan} approach~\cite{isola2017imagetoimage} used in this work, which is a conditional \gls*{gan} able to learn the relation between an image and its label file.
This network is able to generate a variety of image types, which can be employed in various tasks such as photo generation and semantic segmentation. 
Since the results obtained by the \gls*{gan} approach presented some small noise particles, we employed, as a post processing stage, mathematical morphology to remove them, and the results obtained by this method are presented in the Section~\ref{tab:results}.

The \textbf{\gls{encnet}} presented in \cite{zhang2018context} captures the global information of an image to improve scene segmentation.
The model uses the \resnet as feature extractor and uses the extracted features to feed a Context Encoding Module inspired by the Encoding Layer presented in \cite{zhang2017deep}. 

\begin{figure*}[!tb]
\centering
\includegraphics[width=1\linewidth]{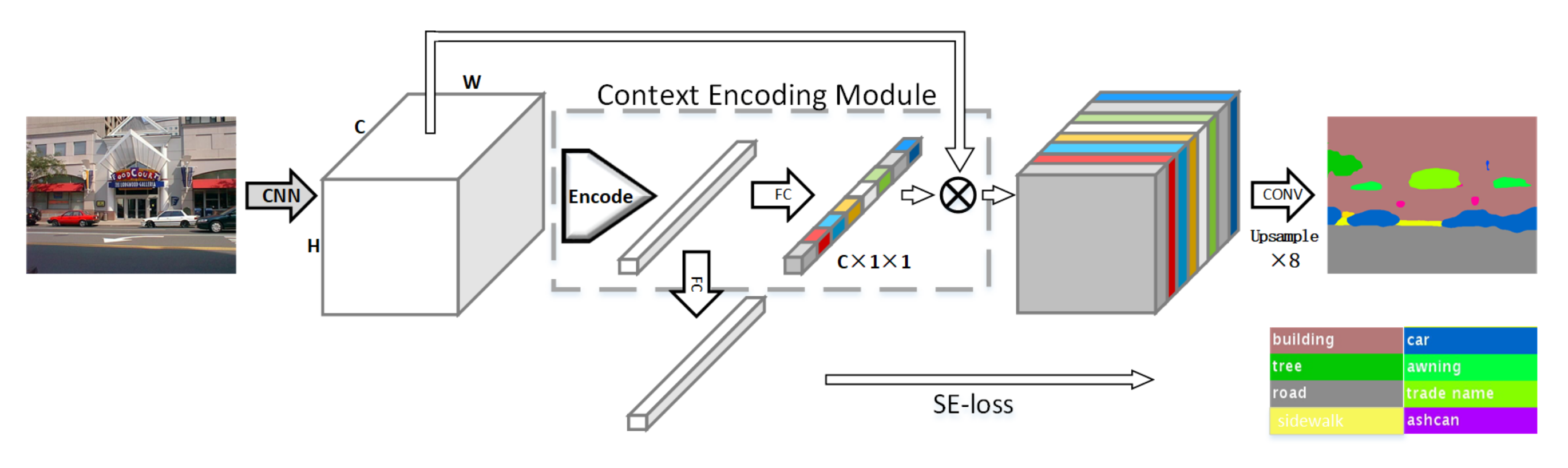}
\caption{\acrshort{encnet} architecture \cite{zhang2018context} employed in the simultaneous ocular region components segmentation.}
\label{fig:encnet}
\end{figure*}

The Context Encoding Module learns visual centers and smoothing factors to create an embedding taking into account the contextual information while highlighting class-dependent feature maps. 
The module is composed of scaling factors to learn the contextual information using fully connected layers, and by a \gls{seloss} to regularize the training of the module by detecting the presence of object classes.
The outputs of the Context Encoding Module are reshaped and processed by a dilated convolution strategy while minimizing the \acrshort{seloss}es and the final pixel-wise loss, as illustrated in Figure \ref{fig:encnet}.

\begin{table*}[!ht]
	\centering
	\caption{Wilcoxon signed-rank results taking into account the sclera segmentation scenario in the \miche dataset.}
	\label{tab:wilcoxon_results}
	\begin{center}
		\resizebox{1\linewidth}{!}{%
			
			{\begin{tabular}{ccccccccccc}
				\toprule
				\multicolumn{1}{l}{}       & \multicolumn{1}{l}{\textbf{RoI}} & \multicolumn{1}{c}{\textbf{EncNet ResNet 50}} & \multicolumn{1}{c}{\textbf{EncNet ResNet 101}} & \multicolumn{1}{c}{\textbf{EncNet ResNet 152}} & \multicolumn{1}{c}{\textbf{FCN}} & \multicolumn{1}{c}{\textbf{GAN}} & \multicolumn{1}{c}{\textbf{GAN PROCESSED}} & \multicolumn{1}{c}{\textbf{OrcNet ResNet 50}} & \multicolumn{1}{c}{\textbf{OrcNet ResNet 101}} & \multicolumn{1}{c}{\textbf{OrcNet ResNet 152}} \\
				\\midrule
                \textbf{EncNet ResNet-50}  & \multirow{9}{*}{\textbf{ALL}}    & -                                             & $1.33^{-3}$                                    & $1.02^{-5}$                                       & $8.38^{-80}$                        & $8.56^{-75}$                        & $7.22^{-66}$                                  & $7.22^{-72}$                                     & $2.61^{-70}$                                      & $3.85^{-67}$                                      \\
            \textbf{EncNet ResNet-101} &                                  & $1.33^{-3}$                                      & -                                              & $3.03^{-10}$                                      & $3.29^{-75}$                        & $6.10^{-64}$                        & $4.34^{-56}$                                  & $1.19^{-71}$                                     & $2.79^{-76}$                                      & $4.61^{-73}$                                      \\
            \textbf{EncNet ResNet-152} &                                  & $1.02^{-5}$                                      & $3.03^{-10}$                                      & -                                              & $1.56^{-80}$                        & $1.00^{-74}$                        & $4.22^{-69}$                                  & $2.24^{-63}$                                     & $1.05^{-64}$                                      & $3.13^{-64}$                                      \\
            \textbf{FCN}               &                                  & $8.38^{-80}$                                     & $3.29^{-75}$                                      & $1.56^{-80}$                                      & -                                & $1.75^{-53}$                        & $9.50^{-67}$                                  & $4.07^{-82}$                                     & $9.58^{-84}$                                      & $2.01^{-81}$                                      \\
            \textbf{GAN}               &                                  & $8.56^{-75}$                                     & $6.10^{-64}$                                      & $1.00^{-74}$                                      & $1.75^{-53}$                        & -                                & $9.78^{-76}$                                  & $5.15^{-76}$                                     & $4.26^{-76}$                                      & $4.38^{-80}$                                      \\
            \textbf{GAN PROCESSED}     &                                  & $7.22^{-66}$                                     & $4.34^{-56}$                                      & $4.22^{-69}$                                      & $9.50^{-67}$                        & $9.78^{-76}$                        & -                                          & $7.91^{-75}$                                     & $6.76^{-75}$                                      & $1.19^{-77}$                                      \\
            \textbf{OrcNet ResNet-50}  &                                  & $7.22^{-72}$                                     & $1.19^{-71}$                                      & $2.24^{-63}$                                      & $4.07^{-82}$                        & $5.15^{-76}$                        & $7.91^{-75}$                                  & -                                             & $9.69^{-3}$                                       & $1.78^{-02}$                                      \\
            \textbf{OrcNet ResNet-101} &                                  & $2.61^{-70}$                                     & $2.79^{-76}$                                      & $1.05^{-64}$                                      & $9.58^{-84}$                        & $4.26^{-76}$                        & $6.76^{-75}$                                  & $9.69^{-3}$                                      & -                                              & $1.29^{-04}$                                      \\
            \textbf{OrcNet ResNet-152} &                                  & $3.85^{-67}$                                     & $4.61^{-73}$                                      & $3.13^{-64}$                                      & $2.01^{-81}$                        & $4.38^{-80}$                        & $1.19^{-77}$                                  & $1.78^{-02}$                                     & $1.29^{-04}$                                      & -                                              \\
            \midrule
            \textbf{EncNet ResNet-50}  & \multirow{9}{*}{\textbf{Iris}}   & -                                             & $8.36^{-3}$                                       & $8.97^{-3}$                                       & $1.58^{-81}$                        & $8.25^{-44}$                        & $6.06^{-47}$                                  & $8.63^{-48}$                                     & $3.14^{-45}$                                      & $9.97^{-47}$                                      \\
            \textbf{EncNet ResNet-101} &                                  & $8.36^{-3}$                                      & -                                              & $6.03^{-3}$                                       & $3.48^{-76}$                        & $3.42^{-36}$                        & $1.66^{-38}$                                  & $6.86^{-35}$                                     & $4.71^{-46}$                                      & $1.30^{-43}$                                      \\
            \textbf{EncNet ResNet-152} &                                  & $8.97^{-3}$                                      & $6.03^{-3}$                                       & -                                              & $9.50^{-83}$                        & $1.17^{-43}$                        & $1.04^{-46}$                                  & $1.90^{-39}$                                     & $4.60^{-47}$                                      & $3.24^{-48}$                                      \\
            \textbf{FCN}               &                                  & $1.58^{-81}$                                     & $3.48^{-76}$                                      & $9.50^{-83}$                                      & -                                & $2.42^{-75}$                        & $1.28^{-72}$                                  & $1.86^{-79}$                                     & $1.14^{-80}$                                      & $2.85^{-79}$                                      \\
            \textbf{GAN}               &                                  & $8.25^{-44}$                                     & $3.42^{-36}$                                      & $1.17^{-43}$                                      & $2.42^{-75}$                        & -                                & $1.87^{-24}$                                  & $1.34^{-65}$                                     & $9.61^{-67}$                                      & $1.02^{-69}$                                      \\
            \textbf{GAN PROCESSED}     &                                  & $6.06^{-47}$                                     & $1.66^{-38}$                                      & $1.04^{-46}$                                      & $1.28^{-72}$                        & $1.87^{-24}$                        & -                                          & $1.56^{-66}$                                     & $1.68^{-67}$                                      & $1.94^{-70}$                                      \\
            \textbf{OrcNet ResNet-50}  &                                  & $8.63^{-48}$                                     & $6.86^{-35}$                                      & $1.90^{-39}$                                      & $1.86^{-79}$                        & $1.34^{-65}$                        & $1.56^{-66}$                                  & -                                             & $2.77^{-1}$                                       & $3.75^{-3}$                                       \\
            \textbf{OrcNet ResNet-101} &                                  & $3.14^{-45}$                                     & $4.71^{-46}$                                      & $4.60^{-47}$                                      & $1.14^{-80}$                        & $9.61^{-67}$                        & $1.68^{-67}$                                  & $2.77^{-1}$                                      & -                                              & $3.49^{-4}$                                       \\
            \textbf{OrcNet ResNet-152} &                                  & $9.97^{-47}$                                     & $1.30^{-43}$                                      & $3.24^{-48}$                                      & $2.85^{-79}$                        & $1.02^{-69}$                        & $1.94^{-70}$                                  & $3.75^{-3}$                                      & $3.49^{-4}$                                       & -                                              \\
            \midrule
            \textbf{EncNet ResNet-50}  & \multirow{9}{*}{\textbf{Sclera}} & -                                             & $9.22^{-03}$                                      & $8.23^{-4}$                                       & $3.12^{-77}$                        & $2.75^{-40}$                        & $1.62^{-32}$                                  & $2.93^{-68}$                                     & $6.67^{-66}$                                      & $2.60^{-72}$                                      \\
            \textbf{EncNet ResNet-101} &                                  & $9.22^{-3}$                                      & -                                              & $1.99^{-5}$                                       & $5.77^{-75}$                        & $2.18^{-31}$                        & $9.22^{-24}$                                  & $6.12^{-67}$                                     & $1.41^{-72}$                                      & $4.23^{-77}$                                      \\
            \textbf{EncNet ResNet-152} &                                  & $8.23^{-4}$                                      & $1.99^{-5}$                                       & -                                              & $2.94^{-77}$                        & $1.67^{-41}$                        & $6.73^{-33}$                                  & $1.00^{-64}$                                     & $1.47^{-68}$                                      & $4.99^{-76}$                                      \\
            \textbf{FCN}               &                                  & $3.12^{-77}$                                     & $5.77^{-75}$                                      & $2.94^{-77}$                                      & -                                & $1.30^{-67}$                        & $2.81^{-71}$                                  & $6.62^{-83}$                                     & $7.03^{-83}$                                      & $4.21^{-83}$                                      \\
            \textbf{GAN}               &                                  & $2.75^{-40}$                                     & $2.18^{-31}$                                      & $1.67^{-41}$                                      & $1.30^{-67}$                        & -                                & $1.74^{-26}$                                  & $1.23^{-74}$                                     & $1.35^{-75}$                                      & $2.33^{-78}$                                      \\
            \textbf{GAN PROCESSED}     &                                  & $1.62^{-32}$                                     & $9.22^{-24}$                                      & $6.73^{-33}$                                      & $2.81^{-71}$                        & $1.74^{-26}$                        & -                                          & $2.06^{-72}$                                     & $6.32^{-73}$                                      & $3.76^{-76}$                                      \\
            \textbf{OrcNet ResNet-50}  &                                  & $2.93^{-68}$                                     & $6.12^{-67}$                                      & $1.00^{-64}$                                      & $6.62^{-83}$                        & $1.23^{-74}$                        & $2.06^{-72}$                                  & -                                             & $4.85^{-3}$                                       & $1.29^{-13}$                                      \\
            \textbf{OrcNet ResNet-101} &                                  & $6.67^{-66}$                                     & $1.41^{-72}$                                      & $1.47^{-68}$                                      & $7.03^{-83}$                        & $1.35^{-75}$                        & $6.32^{-73}$                                  & $4.85^{-3}$                                      & -                                              & $9.48^{-13}$                                      \\
            \textbf{OrcNet ResNet-152} &                                  & $2.60^{-72}$                                     & $4.23^{-77}$                                      & $4.99^{-76}$                                      & $4.21^{-83}$                        & $2.33^{-78}$                        & $3.76^{-76}$                                  & $1.29^{-13}$                                     & $9.48^{-13}$                                      & -                                  \\
				\bottomrule
			\end{tabular}}
			
		}
	\end{center}
\end{table*}


\section{Results and Discussion}
\label{sec:results_and_discussion}

The experiments were carried out using the protocol presented in Section~\ref{subsec:evaluation_protocol}. 
To validate, evaluate and compare the proposed approaches, we report the $F-Score$, the  \glsreset{iou}\gls{iou}, and \glsreset{eer}\gls*{eer}, respectively.

Table~\ref{tab:results_mean} presents the mean results obtained in the performed experiments. 
This table is divided into three sections, one to each segmentation scenario (ALL (Iris and Sclera are considered as a unique region), Iris and Sclera). 
For each one of the scenarios, four segmentation approaches were evaluated, that is, \gls{encnet}, \gls{fcn}, \gls{gan}, and our proposal \gls{ercnet}.
Table~\ref{tab:results} in turn presents the individual result for each \miche database subset for the same scenario division presented in Table~\ref{tab:results_mean}.

Yet, regarding the employed segmentation approaches, it is necessary to highlight that for the \gls{encnet},  \gls{ercnet}, and \gls{gan} approaches, more than one segmentation approach was employed.
For the \gls{encnet} and \gls{ercnet} three segmentation variants were proposed, being the \resnet backbone (\resnet-50, \resnet-101 and \resnet-152) the difference among them. 
For the \gls{gan} approach, two different approaches were employed, one based on the method proposed by~\cite{goodfellow2014generative} and another one in which small noise particles in the output images were removed by mathematical morphology operations.

By analyzing the obtained results presented in Table~\ref{tab:results_mean}, we observed that in all of the segmentation scenarios ---ALL(Iris $+$ Sclera), Iris and Sclera---, the \gls{ercnet} approach presented the best results in terms of  $F-Score$, \glsreset{iou}\gls*{iou}, and \glsreset{eer}\gls*{eer}.
Also, it is important to observe that independently from the used \resnet backbone, the proposed \gls{ercnet} approach outperforms all the baselines evaluated in this work.
Finally, we verify if there was a statistical difference on the obtained results. 
Employing the Wilcoxon signed-rank, we conclude that there is statistical difference between the proposed architecture and the state-of-the-art approaches used as baseline as shown in Table~\ref{tab:wilcoxon_results}.

Once a general analysis of the performance obtained with the \acrshort{ercnet} was presented, we can individually discuss the results obtained in each of the scenarios evaluated using the protocol proposed in Section~\ref{subsec:evaluation_protocol}.  In ALL segmentation scenario, we observed that the mean $F-Score$ obtained by using the \acrshort*{ercnet}~\resnet-152 is $95.67\%$, outperforming the best baseline (\acrshort*{encnet}~\resnet-152) score by $1.42\%$. 
In terms of \acrshort{eer} the obtained score outperforms the best baseline by $33\%$, reducing the error rate of $0$.$39\%$ to $0$.$26\%$.
Finally, by analyzing the \acrshort{iou}, we observed that in the ALL segmentation scenario, the proposed approach achieved a score of $90.97\%$, outperforming the best baseline result by $5.04\%$.



The behaviour presented in the ALL (iris + sclera) segmentation scenario can also be seen in the iris and sclera individual segmentation tasks, since the best results were obtained by using the \gls{ercnet}. 
In the Iris segmentation the mean F-Score, \gls{eer} and \gls{iou} obtained values are  $95.91\%$, $0.15\%$ and $91.17\%$ outperforming the best baseline results by $0.89\%$, $25\%$ and $3.22\%$ respectively. And, in the sclera segmentation the mean  F-Score, \gls{eer} and \gls{iou} obtained values are $89.78\%$, $0.24\%$ and $80.90\%$, outperforming the best results by $4.50\%$, $26.47\%$ and $11.01\%$, respectively.

In addition to presenting better results in relation to the baselines, we also found that the standard deviation presented by the results was smaller when the context based segmentation was employed. 
In this way it is possible to state that the \acrshort{ercnet} presented most stable segmentation of the \glsreset{roi}\gls*{roi}, since it showed less variability in the segmentation masks.
Finally, it is important to highlight that regardless of the backbone used as feature maps extractor in \gls{ercnet}, the proposed approach outperforms all the baselines evaluated.


\begin{table}[!ht]
    
    \centering
    
\caption{Performance comparison among the baseline approaches and the proposed \gls{ercnet} approach employing the \miche dataset as input data.  \glsreset{iou}\gls{iou} is considered as the prior measure ranking methods.}
	\label{tab:results_mean}
	\begin{center}
	    \resizebox{0.6\linewidth}{!}{%

			\begin{tabular}{ccccc}
				\toprule
				\textbf{Approach}      & \textbf{RoI}                     & \textbf{F-Score}     & \textbf{\gls*{eer}}     & \textbf{IoU}         \\
				\midrule
				 \gls{encnet} \resnet-50						& \multirow{10}{*}{ALL}     & $94.21 \pm 4.97$  					 & $0.39 \pm 0.29$ 					 & $86.51 \pm 13.31$  \\
				\gls{encnet} \resnet-101           												&                          							& $94.23 \pm 5.29$  					 & $0.39 \pm 0.29$ 					 & $85.90 \pm 14.51$   \\
				\gls{encnet} \resnet-152           												&                          							& $94.33 \pm 5.54$  					 & $0.39 \pm 0.31$ 					 & $86.60 \pm 13.92$  \\	
				FCN           																&                          								& $83.53 \pm 11.06$ 					 & $1.08 \pm 1.12$ 					 & $72.79 \pm 14.97$  \\
				GAN           																&                          								& $89.80 \pm 8.56$  					 & $0.67 \pm 0.42$ 					 & $82.10 \pm 11.67$  \\
				GAN PROCESSED 																&                          								& $91.04 \pm 9.22$  					 & $0.57 \pm 0.41$ 					 & $84.25 \pm 11.90$  \\
				\gls*{ercnet} \resnet-50  													&                          								& $95.41 \pm 4.22$  					 & $0.27 \pm 0.21$ 					 & $90.30 \pm 10.56$  \\
				\gls*{ercnet} \resnet-101  													&                          								& $95.51 \pm 4.30$  					 & $0.27 \pm 0.21$ 					 & $90.64 \pm 9.92$  \\
				\gls*{ercnet} \resnet-152   													&                          								& $\textbf{95.67} \pm \textbf{4.01}$  & $\textbf{0.26} \pm \textbf{0.22}$ & $\textbf{90.97} \pm \textbf{9.35}$  \\
				\midrule
				\gls{encnet} \resnet-50											    			& \multirow{10}{*}{Iris}   & $94.91 \pm 5.63$ 					 & $0.20 \pm 0.19$ 					 & $88.40 \pm 14.39$  \\
				\gls{encnet} \resnet-101           												&                          & $94.99 \pm 5.69$  				 	 & $0.20 \pm 0.19$ 					 & $87.92 \pm 15.43$  \\
				\gls{encnet} \resnet-152           												&                          & $95.06 \pm 5.39$  				 	 & $0.20 \pm 0.20$ 					 & $88.32 \pm 14.70$  \\	
				FCN           																	&                          & $80.73 \pm 12.19$ 					 & $0.63 \pm 0.62$ 					 & $68.58 \pm 16.43$ \\
				GAN           																	&                          & $88.35 \pm 9.89$ 					 & $0.38 \pm 0.31$ 					 & $79.48 \pm 14.77$ \\
				GAN PROCESSED 																	&                          & $89.14 \pm 9.54$	 				 & $0.36 \pm 0.30$ 					 & $80.26 \pm 15.36$ \\
				\gls*{ercnet} \resnet-50  														&                          & $95.62 \pm 4.81$  					 & $0.16 \pm 0.18$ 					 & $90.42 \pm 12.56$  \\
				\gls*{ercnet} \resnet-101  														&                          & $95.70 \pm 5.46$  					 & $0.16 \pm 0.18$ 					 & $90.82 \pm 11.97$  \\
				\gls*{ercnet} \resnet-152   														&                          & $\textbf{95.91} \pm \textbf{4.96}$  & $\textbf{0.15} \pm \textbf{0.16}$ & $\textbf{91.17} \pm \textbf{11.75}$  \\
				\midrule
				\gls{encnet} \resnet-50															& \multirow{10}{*}{Sclera} & $85.74 \pm 10.15$  				 	 & $0.34 \pm 0.24$ 					 & $73.09 \pm 16.94$  \\
				\gls{encnet} \resnet-101           												&                          & $85.60 \pm 9.79$  				 	 & $0.34 \pm 0.23$ 					 & $72.48 \pm 17.89$  \\
				EncNet \resnet-152           														&                          & $85.91 \pm 10.25$					 & $0.34 \pm 0.26$ 					 & $72.87 \pm 17.93$  \\	
				FCN           																	&                          & $72.41 \pm 15.49$ 					 & $0.74 \pm 0.67$ 					 & $57.35 \pm 17.33$ \\
				GAN           																	&                          & $90.35 \pm 13.17$ 					 & $0.58 \pm 0.36$ 					 & $68.34 \pm 15.21$ \\
				GAN PROCESSED 																	&                          & $82.02 \pm 12.19$ 					 & $0.51 \pm 0.33$ 					 & $70.26 \pm 14.94$ \\
				\gls*{ercnet} \resnet-50  														&                          & $89.42 \pm 7.54$  					 & $0.25 \pm 0.19$ 					 & $80.14 \pm 13.61$  \\
				\gls*{ercnet} \resnet-101  														&                          & $89.46 \pm 7.61$  					 & $0.25 \pm 0.20$ 					 & $80.40 \pm 13.11$  \\
				\gls*{ercnet} \resnet-152   														&                          & $\textbf{89.78} \pm \textbf{6.97}$  & $\textbf{0.25} \pm \textbf{0.20}$ & $\textbf{80.90} \pm \textbf{12.84}$  \\

				\bottomrule
			\end{tabular}
			
		}
	\end{center}
	
\end{table}

\begin{table}[!ht]
	\centering
	\caption{Performance comparison among the baseline approaches and the proposed \gls{ercnet} approach employing the \miche dataset as input data. \glsreset{iou}\gls{iou} is considered as the prior measure ranking methods.} 
	\label{tab:results}
	\begin{center}
		\resizebox{1\linewidth}{!}{%
			
			\begin{tabular}{c|ccccc}
				\toprule
				Dataset & \textbf{Approach}      & \textbf{RoI}                     & \textbf{F-Score}     & \textbf{\gls*{eer}}     & \textbf{IoU}         \\
				\midrule
				\multirow{30}{*}{\rotatebox[origin=c]{90}{\acrshort*{gs4}}} & \gls{encnet} \resnet-50						& \multirow{10}{*}{ALL}    & $93.77 \pm 5.31$  					 & $0.34 \pm 0.22$ 					 & $ 86.31 \pm 11.81$  \\
				& \gls{encnet} \resnet-101           												&                          & $93.40 \pm 5.41$  					 & $0.40 \pm 0.24$ 					 & $83.80 \pm 13.41$   \\
				& \gls{encnet} \resnet-152           												&                          & $93.51 \pm 4.95$  					 & $0.37 \pm 0.21$ 					 & $85.17 \pm 12.82$  \\	
				& FCN           																&                          & $78.60 \pm 13.19$ 					 & $1.22 \pm 0.92$ 					 & $66.38 \pm 15.36$  \\
				& GAN           																&                          & $92.99 \pm 7.20$  					 & $0.37 \pm 0.23$ 					 & $87.51 \pm 9.75$  \\
				& GAN PROCESSED 																&                          & $93.60 \pm 6.13$  					 & $0.34 \pm 0.22$ 					 & $88.39 \pm 8.53$  \\ 
				& \gls*{ercnet} \resnet-50  													&                          & $94.65 \pm 3.65$  					 & $0.28 \pm 0.17$ 					 & $88.77 \pm 9.26$  \\
				& \gls*{ercnet} \resnet-101  													&                          & $94.78 \pm 3.47$  					 & $0.28 \pm 0.16$ 					 & $89.31 \pm 7.47$  \\
				& \gls*{ercnet} \resnet-152   													&                          & $\textbf{95.09} \pm \textbf{3.32}$  & $\textbf{0.27} \pm \textbf{0.16}$ & $\textbf{89.91} \pm \textbf{6.90}$  \\ \\
				& \gls{encnet} \resnet-50											    			& \multirow{10}{*}{Iris}   & $94.67 \pm 3.70$ 					 & $0.18 \pm 0.13$ 					 & $88.36 \pm 11.65$  \\
				& \gls{encnet} \resnet-101           												&                          & $94.15 \pm 4.82$  				 	 & $0.20 \pm 0.13$ 					 & $86.79 \pm 13.04$  \\
				& \gls{encnet} \resnet-152           												&                          & $94.66 \pm 3.57$  				 	 & $0.19 \pm 0.13$ 					 & $87.51 \pm 11.84$  \\	
				& FCN           																&                          & $69.10 \pm 15.67$ 					 & $0.73 \pm 0.55$ 					 & $52.98 \pm 16.50$ \\
				& GAN           																&                          & $80.12 \pm 11.90$ 					 & $0.51 \pm 0.25$ 					 & $68.13 \pm 13.31$ \\
				& GAN PROCESSED 																&                          & $82.34 \pm 11.52$	 				 & $0.42 \pm 0.22$ 					 & $71.25 \pm 13.15$ \\
				& \gls*{ercnet} \resnet-50  													&                          & $94.81 \pm 3.28$  					 & $0.16 \pm 0.11$ 					 & $89.39 \pm 9.53$  \\
				& \gls*{ercnet} \resnet-101  													&                          & $94.98 \pm 4.08$  					 & $0.16 \pm 0.10$ 					 & $89.98 \pm 8.15$  \\
				& \gls*{ercnet} \resnet-152   													&                          & $\textbf{95.33} \pm \textbf{2.73}$  & $\textbf{0.15} \pm \textbf{0.09}$ & $\textbf{90.62} \pm \textbf{7.58}$  \\ \\
				& \gls{encnet} \resnet-50															& \multirow{10}{*}{Sclera} & $85.52 \pm 8.91$  				 	 & $0.30 \pm 0.21$ 					 & $73.09 \pm 15.23$  \\
				& \gls{encnet} \resnet-101           												&                          & $83.67 \pm 9.64$  				 	 & $0.33 \pm 0.\gls{encnet}$ 					 & $69.70 \pm 16.80$  \\
				& EncNet \resnet-152           												&                          & $85.17 \pm 9.53$					 & $0.32 \pm 0.19$ 					 & $70.90 \pm 17.86$  \\	
				& FCN           																&                          & $69.10 \pm 15.67$ 					 & $0.73 \pm 0.55$ 					 & $52.98 \pm 16.50$ \\
				& GAN           																&                          & $80.12 \pm 11.90$ 					 & $0.51 \pm 0.25$ 					 & $68.13 \pm 13.31$ \\
				& GAN PROCESSED 																&                          & $82.34 \pm 11.52$ 					 & $0.42 \pm 0.22$ 					 & $71.25 \pm 13.15$ \\
				& \gls*{ercnet} \resnet-50  													&                          & $88.27 \pm 6.75$  					 & $0.25 \pm 0.16$ 					 & $78.42 \pm 11.60$  \\
				& \gls*{ercnet} \resnet-101  													&                          & $88.44 \pm 5.77$  					 & $0.25 \pm 0.17$ 					 & $79.02 \pm 9.81$  \\
				& \gls*{ercnet} \resnet-152   													&                          & $\textbf{88.57} \pm \textbf{5.60}$  & $\textbf{0.24} \pm \textbf{0.17}$ & $\textbf{79.51} \pm \textbf{9.41}$  \\
				
				\midrule 
				
				\multirow{30}{*}{\rotatebox[origin=c]{90}{\acrshort*{gt2}}} & \gls{encnet} \resnet-50						& \multirow{10}{*}{ALL}    & $93.57 \pm 5.63$  					 & $0.57 \pm 0.39$ 					 & $83.85 \pm 15.56$  \\
				& \gls{encnet} \resnet-101           													&                          & $94.03 \pm 6.34$  					 & $0.51 \pm 0.39$ 					 & $85.38 \pm 15.97$   \\
				& \gls{encnet} \resnet-152           													&                          & $94.01 \pm 5.83$  					 & $0.54 \pm 0.46$ 					 & $84.75 \pm 16.98$  \\	
				& FCN           																&                          & $89.56 \pm 7.41$ 					 & $0.96 \pm 1.04$ 					 & $81.02 \pm 12.84$  \\
				& GAN           																&                          & $86.31 \pm 10.80$ 					 & $1.12 \pm 0.68$ 					 & $76.72 \pm 13.51$  \\
				& GAN PROCESSED 																&                          & $88.25 \pm 13.19$ 					 & $0.90 \pm 0.65$ 					 & $80.31 \pm 15.71$  \\
				& \gls*{ercnet} \resnet-50  													&                          & $95.19 \pm 4.76$  					 & $0.35 \pm 0.25$ 					 & $89.92 \pm 11.32$  \\
				& \gls*{ercnet} \resnet-101  													&                          & $95.40 \pm 4.65$  					 & $0.34 \pm 0.25$ 					 & $90.30 \pm 11.52$  \\
				& \gls*{ercnet} \resnet-152   													&                          & $\textbf{95.38} \pm \textbf{4.92}$  & $\textbf{0.34} \pm \textbf{0.27}$ & $\textbf{90.37} \pm \textbf{11.32}$  \\ \\
				& \gls{encnet} \resnet-50											    				& \multirow{10}{*}{Iris}   & $94.17 \pm 6.85$ 					 & $0.28 \pm 0.24$ 					 & $86.55 \pm 16.76$  \\
				& \gls{encnet} \resnet-101           													&                          & $94.84 \pm 5.86$  				 	 & $0.26 \pm 0.25$ 					 & $87.39 \pm 17.10$  \\
				& \gls{encnet} \resnet-152           													&                          & $94.79 \pm 6.09$  				 	 & $0.27 \pm 0.26$ 					 & $87.10 \pm 18.08$  \\	
				& FCN           																&                          & $89.93 \pm 7.08$ 					 & $0.56 \pm 0.52$ 					 & $80.80 \pm 14.41$ \\
				& GAN           																&                          & $91.40 \pm 9.59$ 					 & $0.44 \pm 0.41$ 					 & $82.92 \pm 17.39$ \\
				& GAN PROCESSED 																&                          & $91.27 \pm 10.32$	 				 & $0.45 \pm 0.43$ 					 & $82.45 \pm 18.63$ \\
				& \gls*{ercnet} \resnet-50  													&                          & $95.50 \pm 4.68$  					 & $0.21 \pm 0.21$ 					 & $90.14 \pm 13.54$  \\
				& \gls*{ercnet} \resnet-101  													&                          & $95.75 \pm 5.27$  					 & $0.20 \pm 0.25$ 					 & $90.56 \pm 13.92$  \\
				& \gls*{ercnet} \resnet-152   													&                          & $\textbf{95.96} \pm \textbf{4.86}$  & $\textbf{0.19} \pm \textbf{0.20}$ & $\textbf{90.75} \pm \textbf{13.74}$  \\ \\
				& \gls{encnet} \resnet-50																& \multirow{10}{*}{Sclera} & $83.04 \pm 13.10$  				 & $0.48 \pm 0.31$ 					 & $68.98 \pm 18.46$  \\
				& \gls{encnet} \resnet-101           													&                          & $84.67 \pm 12.04$ 				 	 & $0.45 \pm 0.30$ 					 & $71.40 \pm 18.43$  \\
				& \gls{encnet} \resnet-152           													&                          & $83.81 \pm 13.03$					 & $0.47 \pm 0.37$ 					 & $70.37 \pm 19.44$  \\	
				& FCN           																&                          & $76.58 \pm 14.07$ 					 & $0.79 \pm 0.68$ 					 & $62.64 \pm 16.75$ \\
				& GAN           																&                          & $77.92 \pm 14.83$ 					 & $0.81 \pm 0.50$ 					 & $64.97 \pm 16.88$ \\
				& GAN PROCESSED 																&                          & $79.28 \pm 14.41$ 					 & $0.74 \pm 0.45$ 					 & $66.34 \pm 16.90$ \\
				& \gls*{ercnet} \resnet-50  													&                          & $88.37 \pm 10.07$ 					 & $0.32 \pm 0.22$ 					 & $79.51 \pm 14.49$  \\
				& \gls*{ercnet} \resnet-101  													&                          & $88.44 \pm 10.76$  				 & $0.32 \pm 0.26$ 					 & $79.63 \pm 14.84$  \\
				& \gls*{ercnet} \resnet-152   													&                          & $\textbf{88.85} \pm \textbf{10.03}$ & $\textbf{0.32} \pm \textbf{0.24}$ & $\textbf{79.68} \pm \textbf{15.40}$  \\
				\midrule 
				\multirow{30}{*}{\rotatebox[origin=c]{90}{\acrshort*{ip5}}} & \gls{encnet} \resnet-50						& \multirow{10}{*}{ALL}    & $95.29 \pm 3.96$  					 & $0.25 \pm 0.25$ 					 & $89.38 \pm 12.57$  \\
				& \gls{encnet} \resnet-101           													&                          & $95.25 \pm 4.11$  					 & $0.26 \pm 0.24$ 					 & $88.51 \pm 14.15$   \\
				& \gls{encnet} \resnet-152           													&                          & $95.48 \pm 5.83$  					 & $0.25 \pm 0.26$ 					 & $89.89 \pm 11.97$  \\	
				& FCN           																&                          & $82.42 \pm 12.57$ 					 & $1.06 \pm 1.39$ 					 & $70.98 \pm 16.70$  \\
				& GAN           																&                          & $90.09 \pm 7.68$  					 & $0.53 \pm 0.36$ 					 & $82.06 \pm 11.75$  \\
				& GAN PROCESSED 																&                          & $91.27 \pm 8.34$  					 & $0.46 \pm 0.36$ 					 & $84.06 \pm 11.46$  \\
				& \gls*{ercnet} \resnet-50  													&                          & $96.39 \pm 4.24$  					 & $0.19 \pm 0.22$ 					 & $92.21 \pm 11.11$  \\
				& \gls*{ercnet} \resnet-101  													&                          & $96.36 \pm 4.77$  					 & $0.19 \pm 0.21$ 					 & $92.32 \pm 10.76$  \\
				& \gls*{ercnet} \resnet-152   													&                          & $\textbf{96.53} \pm \textbf{3.80}$  & $\textbf{0.18} \pm \textbf{0.23}$ & $\textbf{92.62} \pm \textbf{9.83}$  \\ \\
				& \gls{encnet} \resnet-50											    				& \multirow{10}{*}{Iris}   & $95.91 \pm 6.35$ 					 & $0.14 \pm 0.21$ 					 & $90.28 \pm 14.75$  \\
				& \gls{encnet} \resnet-101           													&                          & $95.99 \pm 6.40$  				 	 & $0.14 \pm 0.20$ 					 & $89.58 \pm 16.15$  \\
				& \gls{encnet} \resnet-152           													&                          & $95.74 \pm 6.51$  				 	 & $0.14 \pm 0.20$ 					 & $90.34 \pm 14.17$  \\	
				& FCN           																&                          & $83.16 \pm 13.83$ 					 & $0.60 \pm 0.79$ 					 & $71.95 \pm 18.37$ \\
				& GAN           																&                          & $93.53 \pm 8.17$ 					 & $0.20 \pm 0.27$ 					 & $87.39 \pm 13.60$ \\
				& GAN PROCESSED 																&                          & $93.81 \pm 6.79$	 				 & $0.21 \pm 0.26$ 					 & $87.07 \pm 14.31$ \\
				& \gls*{ercnet} \resnet-50  													&                          & $96.55 \pm 6.48$  					 & $0.12 \pm 0.21$ 					 & $91.74 \pm 14.61$  \\
				& \gls*{ercnet} \resnet-101  													&                          & $96.38 \pm 7.03$  					 & $0.12 \pm 0.20$ 					 & $91.93 \pm 13.83$  \\
				& \gls*{ercnet} \resnet-152   													&                          & $\textbf{96.45} \pm \textbf{7.30}$  & $\textbf{0.12} \pm \textbf{0.20}$ & $\textbf{92.14} \pm \textbf{13.94}$  \\ \\
				& \gls{encnet} \resnet-50																& \multirow{10}{*}{Sclera} & $88.67 \pm 08.43$  				 & $0.24 \pm 0.21$ 					 & $77.21 \pm 17.12$  \\
				& \gls{encnet} \resnet-101           													&                          & $88.45 \pm 7.70$ 				 	 & $0.25 \pm 0.21$ 					 & $76.34 \pm 18.43$  \\
				& \gls{encnet} \resnet-152           													&                          & $88.74 \pm 8.19$					 & $0.24 \pm 0.23$ 					 & $77.35 \pm 16.49$  \\	
				& FCN           																&                          & $71.54 \pm 16.74$ 					 & $0.69 \pm 0.77$ 					 & $56.44 \pm 18.73$ \\
				& GAN           																&                          & $83.00 \pm 12.77$ 					 & $0.41 \pm 0.34$ 					 & $71.92 \pm 15.45$ \\
				& GAN PROCESSED 																&                          & $84.54 \pm 10.63$ 					 & $0.38 \pm 0.32$ 					 & $73.18 \pm 14.77$ \\
				& \gls*{ercnet} \resnet-50  													&                          & $91.62 \pm 5.79$ 					 & $0.19 \pm 0.18$ 					 & $82.49 \pm 14.75$  \\
				& \gls*{ercnet} \resnet-101  													&                          & $91.49 \pm 6.30$ 	 				 & $0.19 \pm 0.18$ 					 & $82.56 \pm 14.67$  \\
				& \gls*{ercnet} \resnet-152   													&                          & $\textbf{91.93} \pm \textbf{5.27}$  & $\textbf{0.18} \pm \textbf{0.19}$ & $\textbf{83.51} \pm \textbf{13.71}$  \\
				
				\bottomrule
			\end{tabular}
			
		}
	\end{center}
\end{table}

\subsection{Visual  Analysis}

Here, we perform a visual analysis of some segmented images by the baselines and the proposed architecture. 
Fig.~\ref{fig:general_results_comparison} shows segmentation samples of the three evaluated scenarios (\textit{\textbf{ALL}}, \textbf{\textit{iris}} and \textbf{\textit{sclera}}) performed by the \gls{ercnet} approach, as well as the state-of-the-art segmentation approaches employed as baselines.

By analyzing the segmentation results presented in Fig.~\ref{fig:general_results_comparison}, it is possible to observe that the results obtained using the \acrshort{ercnet} approach was slightly better if compared with the baseline approaches.
In all the evaluated segmentation scenarios (ALL(Iris + Sclera), iris and sclera), the proposed approach with a \resnet-152 backbone, performed a better segmentation since fewer pixels were wrongly segmented (green area) while the number of pixels that have not been targeted is also less than that presented by other approaches (red area).

\section{Conclusion}
\label{sec:conclusion}

This work introduced the \glsreset{ercnet}\gls{ercnet}, a new approach that uses the \glsreset{pcloss}\gls*{pcloss} to simultaneously segment the ocular region components using the contextual information.
PC-Loss was implemented, evaluated and compared with baseline methods (\gls{encnet}, \gls{fcn} and \gls{gan}).
The proposed approach attained better \glsreset{iou}\gls{iou}, Error Rate (ER) and $F-Score$ values in all the evaluated scenarios (\textbf{\textit{general}}, \textbf{\textit{iris}} and \textbf{\textit{sclera}} segmentations).

Taking into account the presented results it is necessary to emphasize that the only difference between the \acrshort{encnet} and the \acrshort{ercnet} segmentation approaches is the use of the \acrshort{pcloss}. By using the \acrshort{pcloss}, we observe that the use of contextual information improve the results of a simultaneous iris and sclera segmentation independent of the architecture used. Once it was found that the use of contextual information improve the segmentation results, we prove the hypothesis of this work and answer the main question with ``it is possible to improve the ocular region components extraction by analyzing the context present in the images''.

Additionally, we also manually labeled $3,191$ images from the \miche dataset for sclera and iris segmentation. 
These masks were made available to the research community, enabling other researchers to fairly compare their proposals on the same tasks and also among published works.

There is still room for improvements in the simultaneous segmentation of iris and ocular region research field. 
As future work, we intend to:
(i)~design new and better network architectures; 
(ii)~design a general and independent sensor approach, where the image sensor is firstly classified and then the iris and the ocular region are simultaneously segmented with a specific approach trained for a particular image sensor;
(iii)~compare the proposed context based segmentation approach with methods applied in other domains; 
(iv)~design a new multi-task approach for simultaneously detecting and segmenting the components of the ocular region, taking into account the contextual information of the images.

\bibliographystyle{unsrt}  
\bibliography{references}

\end{document}